\documentclass{egpubl}
\usepackage{3dor20}
\usepackage{adjustbox}
\usepackage{amsmath}
\usepackage{amssymb}
\usepackage{subfigure}
\usepackage{latexsym}
\usepackage[mathscr]{euscript}
\usepackage{wasysym}
\usepackage{url}
\usepackage{framed,multirow}

\ConferenceSubmission   %
\WsPaper           %

\usepackage[T1]{fontenc}
\usepackage{dfadobe}  

\usepackage{cite}  %
\BibtexOrBiblatex
\electronicVersion
\PrintedOrElectronic

\usepackage{egweblnk}

\title[SHREC 2020 track:  6D object pose estimation]%
{SHREC 2020 track:  6D object pose estimation}

\author[H. Yuan et al.]
{\parbox{\textwidth}{\centering Honglin  Yuan$^{1,*}$,
		Remco  C. Veltkamp$^{1,*}$,  
		Georgios  Albanis$^2$, 
		Nikolaos  Zioulis$^2$, 
		Dimitrios  Zarpalas$^2$, 
		Petros  Daras$^2$
	}
	\\
	{\parbox{\textwidth}{\centering $^*$ Track organizers\\$^1$Utrecht University, Netherlands\\
			$^2$Centre for Research and Technology  Hellas, Greece\\
		}
	}
}

\begin{document}

	\maketitle
	\begin{abstract}
		6D pose estimation is crucial for augmented reality, virtual reality, robotic  manipulation and visual navigation. However, the problem is challenging due to the variety of objects in the real world. They have varying 3D shape and  their appearances in captured images  are affected by sensor noise, changing lighting conditions and occlusions between objects.  Different  pose estimation methods have different strengths and weaknesses, depending on feature representations and scene contents.  At the same time,  existing 3D datasets that are used for data-driven methods to estimate 6D poses have limited view angles and low resolution. 
		
		To address these issues, we organize the Shape Retrieval Challenge benchmark on 6D pose estimation and   create a physically accurate simulator that is able to generate photo-realistic color-and-depth image pairs with  corresponding ground truth 6D poses. From captured color and depth images,  we use this simulator to generate a 3D dataset which has  400 photo-realistic synthesized color-and-depth image pairs with various view angles for training, and  another 100 captured and synthetic images for testing. Five research groups register in this track and two of them  submitted their results. %
		
		Data-driven methods are the current trend in 6D object pose estimation and our evaluation results show that approaches which fully exploit the color and geometric features are more robust  for  6D pose estimation of reflective and texture-less objects  and occlusion. This   benchmark and  comparative evaluation results  have the potential to further enrich and boost the research of 6D object pose estimation and its applications.

	\end{abstract}  

\section{Introduction}

The ability to estimate 6D object pose including its orientation  and location  is essential for many applications, such as visual navigation, robot manipulation and virtual reality.  The awareness of the 3D rotation and 3D translation matrix of objects in a scene is  referred to as 6D, where the D stands for degrees of freedom pose. While it is possible to obtain the 6D pose with hand-crafted features \cite{mur2015orb}, these methods fail to predict poses for texture-less objects. With the advent of cheap RGB-D sensors, the precision of 6D object pose estimation is improved for both rich  and low texture objects \cite{tekin2018real}. Nonetheless, it remains a challenge as  accurate 6D object pose  and real-time object instance recognition are both required for the real-world applications.

Traditional 6D object pose estimation approaches work by first extracting color features from the RGB image and performing feature matching to get correspondences. Based on these correspondences, the 6D pose is estimated by solving a Perspective-n-Point (PnP) problem \cite{kneip2014upnp}. 
Hand-crafted features, such as SIFT \cite{ng2003sift} and ORB \cite{mur2015orb}, are often used   by these methods, for they  are  robust to  scale, rotation, illumination and  view angles. However, the heavy dependence  on hand-crafted features and fixed matching process have limited  empirical performances of these methods to predict 6D poses for texture-less object in poor light conditions   or   clustered scenes.

The emergency of commodity depth cameras  has enabled many  methods with RGB-D images as input \cite{choi2016rgb,zhang2017texture} to estimate more accurate 6D pose for texture-less objects.  Choi et al. \cite{choi2016rgb} introduce a voting-based approach which further incorporates geometric  and  color information to predict poses in clustered scenes. To handle  low texture objects, Hinterstoisser et al. \cite{hinterstoisser2011multimodal} propose template matching approach that builds  different modalities to detect the known object and then estimate 6D poses. However, template-based methods are not robust to changing lighting conditions and occlusions.  To address these issues, Brachmann et al.  \cite{brachmann2014learning}   first regress an intermediate  object coordinate with different voting scores which are used to  predict  correspondences and then  predict the object pose with these correspondences.

More recently, Convolutional Neural Networks (CNNs) and deep learning have been applied to the 6D pose estimation problem. PoseCNN \cite{xiang2017posecnn} and PoseNet \cite{kendall2015posenet} directly regress from a  RGB image to a 6D object pose by a CNN-based architecture. Unlike PoseCNN,  \cite{tremblay2018deep,pavlakos20176} first predicts 2D projection of predefined 3D key points and then use these correspondence to estimate poses. The aforementioned approaches do not utilize the depth information,  resulting in  failing  to predict  poses for the same object with different scales.

To further improve the performance of the 6D object pose estimation, recent approaches \cite{xiang2017posecnn,jafari2018ipose,wang2019densefusion} combine the  color and depth information   to estimate the 6D object pose.  Ipose \cite{jafari2018ipose}  first processes the RGB  image with the encoder-decoder architecture to obtain a coarse 6D pose and then refines the pose by the iterative closest point (ICP) algorithm based on depth information. However, the refinement method using ICP is time-consuming and cannot achieve real-time inference speed. Instead of using color and depth information separately, Michel et al. \cite{michel2017global} fuse the color and depth information in the early stage, where the  depth information is treated as a fourth channel and concatenated with RGB channels. Other solutions including Densefusion \cite{wang2019densefusion,cheng20196d},  fuse the depth information in the later stage, which first extract visual and geometric features and then fuse  these features together. The fused features are used to directly predict the 6D object pose. With depth information, these methods are more robust to occlusion and changing light conditions.

Even though more and more algorithms, aiming to estimate the 6D object pose  have been published, it is unclear how well scenarios and methods perform. New approaches are usually compared with only a few competitors on a  particular dataset. To address these issues,  the BOP benchmark \cite{hodan2018bop} is proposed, which combines eight datasets in a unified format.  However, their datasts    have several limitations: the objects are often located in the center of  the image plane; images are generated  in similar distances; generating these datasets has  high cost (time and money) associated with
ground truth annotation. Since we use synthetic data, we can   provide high-quality data with minimum cost (e.g., human labor ). To compare and evaluate algorithms for robotics grasping, the OpenGRASP benchmarking suite \cite{ulbrich2011opengrasp}  provides the simulation environment  containing test cases, robot models and scenarios to test methods and rank them.  However, the simulation environment is not photo-realistic and has the reality gap, while our simulator provides high-resolution extremely realistic images.

 The LineMOD \cite{hinterstoisser2011multimodal} and YCB-Video \cite{ xiang2017posecnn} dataset  are the two mostly used 3D object datasets for the 6D pose estimation. However,  the view angles of captured images are limited and the objects are not easily accessible to other researchers. Other works \cite{dosovitskiy2015flownet} combine real and synthesized data to generate 3D object datasets, which render 3D object models on real backgrounds to produce images.   While the backgrounds are realistic, the synthesized images are not photo-realistic. For example,  the rendered objects are flying midair and out of context \cite{dosovitskiy2015flownet}.  Unlike these approaches, we use depth image based rendering (DIBR)  to generate  a 3D object dataset, which provides photo-realistic color-and-depth image pairs with ground truth 6D poses.

Our main contributions are summarized as follows:

(1) Datasets.  Our training dataset is generated by a free-viewpoint DIBR  approach, which provides a large amount of high-resolution  photo-realistic color-and-depth images pairs with ground truth 6D poses. Besides, the synthesized images have  plausible physical locations, lighting, and scale.  A testing dataset combines real captured and synthesized images for testing approaches. Our datasets contain 3D models  with a wide range of sizes, shapes, texture and occlusion.

(2)  A comprehensive evaluation of 6D object pose estimation approaches. We organize the Shape Retrieval Challenge (SHREC) benchmark on 6D pose estimation and use different evaluation metrics to compare the proposed methods based on our datasets.  Evaluation results indicate that approaches that fully exploit the color and geometric features are more robust  for  6D pose estimation of reflective and texture-less objects  and occlusion.

\section{Benchmark}
Our 6D object pose estimation retrieval benchmark includes $8$ objects of varying shape and texture. It has high resolution color-and-depth image pairs and high-quality 3D models. To  facilitate data-driven approaches, we apply DIBR to generate 400 synthesized color-and-depth image pairs with   resolution of $1280 \times 720$ for training.  Another $100$  captured and synthesized  color and depth images are used for testing.

	\subsection{Dataset}
\begin{figure}[htb]
	\centering
	\includegraphics[width=0.48\textwidth]{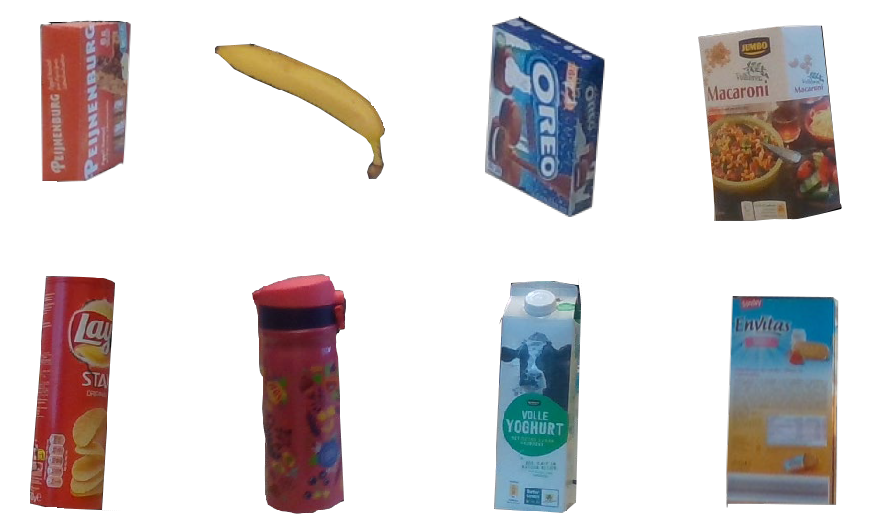}
	\caption{Overview of the dataset}
	\label{Fig.1}
\end{figure}

In order to cover as many aspects of pose estimation challenges as possible, our dataset contains a variety of objects with different sizes, shapes, texture, and reflective characteristics. For example, it is a challenge to estimate the pose of the texture-less object. Thus, when selecting objects, this issue should be considered.  Besides, we also consider the portability. We aim to provide  datasets  with easily carrying, shipping and stored  objects. In order to make the dataset reproducible,  the cost of the object is taken into consideration as well. We choose the   popular consumer products  which are low price and easy to buy.  With consideration of   these practical issues,   we choose eight representative objects to  create our dataset, as shown in Figure \ref{Fig.1}. 

We use both real-world data and data generated from simulation for the 6D object pose estimation retrieval.  The real-world data is captured by  the Intel RealSense depth   camera D415 with  resolution of $1280 \times 720$.  However, the captured color-and-depth image pairs suffer from motion blur and misalignment  caused by the hand-hold camera. We  remove blurring images with the blur metric \cite{crete2007blur}. Even though an alignment method from Intel RealSense camera is applied, the captured color and depth image are still misaligned, especially when the camera is near the object. This is because it is difficult for previous approaches to find correspondences between color and depth images to achieve alignment. 

Instead of finding correspondences between color-and-depth image pairs, we create a different depth map for each color image, which has better alignment with the color image. This map is generated by  multi-view stereo from COLMAP \cite{schonberger2016structure}, a state-of-the-art 3D reconstruction system. Then we align the captured depth map to the estimated map by comparing the values and normals between  two maps.  In this way, we align the captured depth map to the captured color image. Apart from depth maps, we  estimate  object poses using the Structure-from-Motion(SfM)  from COLMAP.
\begin{figure}[htb]
	\centering
	\includegraphics[scale=0.5]{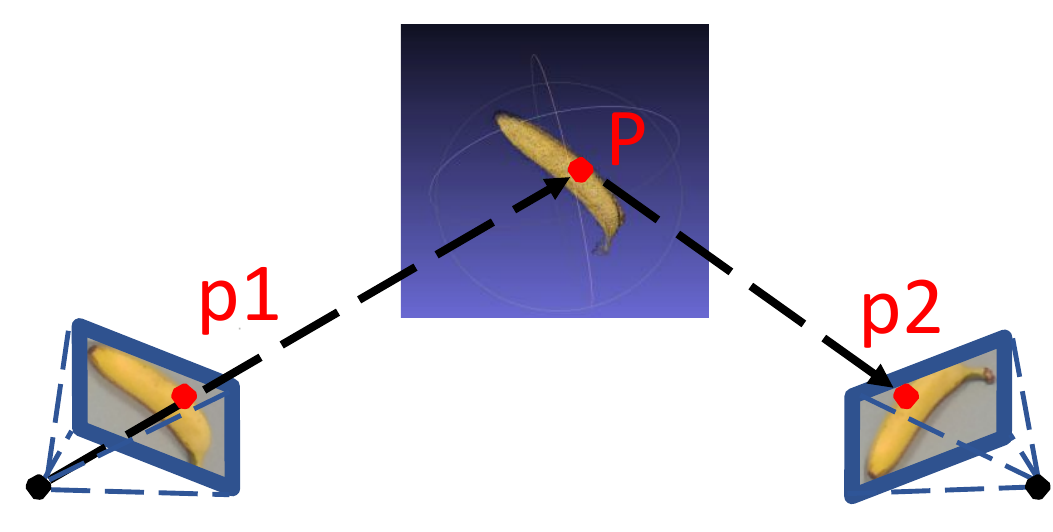}
	\caption{The 3D warping process. It  projects a point  $p_{1}$ in the left image plane  to a world point $P$ and then   $P$ is projected into   the pixel position $p_{2}$ in the right image plane.}
	\label{Fig.4}
\end{figure}

Our aim is to build  3D datasets containing rich  viewpoints, scales and high resolutions which are the limitation for captured datasets \cite{hinterstoisser2011multimodal,xiang2017posecnn}. Inspired by the low cost of producing very large-scale synthetic datasets with complete and accurate ground-truth information, as well as the recent successes of synthetic data
for training 6D pose estimation systems, we build  a physically accurate simulator. Our simulator is  based on DIBR and  can generate high resolution photot-realistic color and depth images, and their corresponding ground truth 6D poses. DIBR performs 3D warping that projects pixels in the reference image  to the  world coordinate and then projects the world points  to  new positions in another image plane to get the new image, which avoids the global 3D reconstruction of the scene. Figure \ref{Fig.4} shows the 3D warping  process of DIBR.

During simulation,  we provide sufficient variations of viewpoints  to mimic a variety of object locations. For it is difficult to generate new images  from the same image distribution, previous methods often randomly project objects into an arbitrary scene to produce the synthesized image. However, such synthesized images are unrealistic  compared to real-world scenes. On contrast, our simulator provides  realistic imagery with the corresponding 6D pose. 

 We combine the synthesized and captured data to build the  final  dataset. It consists of two subsets: a training set of $400$ synthesized color-and-depth image pairs  and a testing set of $100$ captured and synthesized color-and-depth image pairs. For each  frame we provide the following data:

$\bullet$  6D poses for each object.

$\bullet$ Color images with  resolution of $1280 \times 720$ in PNG.

$\bullet$ Depth images with  resolution of $1280 \times 720$ in PNG.

$\bullet$ Binary segmentation masks for each image.

$\bullet$ 2D bounding boxes for each object.

$\bullet$ 3D point clouds with RGB color and normals for each object.

$\bullet$ Calibration information for each image.

\subsection{Evaluation metrics}

In this benchmark, we require the participants to submit the estimated 6D object poses of the testing set. The performance of  6D object pose estimation  is evaluated by  ADD(-S) which are  the average distance metric (ADD) \cite{hinterstoisser2012model}  and the average closest point distance  (ADD-S)    \cite{xiang2017posecnn}. 

Given the ground truth  rotation matrix $R$ and   translation matrix $T$ and its corresponding  estimated rotation matrix $\hat{R}$ and translation matirx $\hat{T}$, the ADD  computes  mean distances between all 3D model points $x$ transformed by  $[\hat{R}|\hat{T}]$ and $[R|T]$:
\begin{equation}
ADD = \frac{1}{N} \sum_{x\in S } ||(Rx + T) -(\hat{R}x + \hat{T})  ||,
\end{equation}
where $S$ is the set of 3D model points and $N$ is the number of points. 

The ADD-S is an ambiguity-invariant pose error metric, which takes care of both symmetric and non-symmetric objects into an overall evaluation.
\begin{equation}
ADD{\text -}S = \frac{1}{N} \sum_{x_{1}\in S } \min_{x_{2}\in S}||(Rx_{1} + T) -(\hat{R}x_{2} + \hat{T})  ||
\end{equation}
The area under the accuracy-threshold curve (AUC) which is calculated from ADD(-S) is another evaluation metric. Specifically,   if the ADD(-S) is smaller than a threshold defined from the diameter of the 3D object model, we consider the estimated pose is correct. Based on  that, we define a variable range of  thresholds from $0\%$ to $100\%$ of the 3D object diameter  and then compute the ADD(-S) for each threshold. With the two sets of values, we can get the  accuracy-threshold curve referred to as AUC. Then   the area under the AUC is calculated. 

We also use the reprojection error, which is often used for 6D object pose estimation of feature matching methods, as our fourth performance metric. Rather than computing distance between two 3D point pairs, the  reprojection error is calculated by first projecting 3D points into an image plane and then computing the pairwise distances in the image space.

\section{Methods}

All the proposed methods  are described in the following subsections. We  choose  DenseFusion \cite{wang2019densefusion}  as a baseline approach for the 6D object pose estimation.  Two research groups contributed their methods in this joint experimental comparison. 

\subsection{Baseline: DenseFusion: 6D Object Pose Estimation by Iterative Dense Fusion}

DenseFusion \cite{wang2019densefusion} is a heterogeneous neural network architecture with RGB-D images as input. It processes color and depth information separately and  uses a dense fusion network to extract pixel-wise dense features, from which the 6D object pose is estimated. Furthermore, an end-to-end iterative pose refinement network is proposed to further improve the accuracy of the predicted pose while achieving  real-time speed.

\begin{figure}[htb]
	\centering
	\includegraphics[scale=0.3]{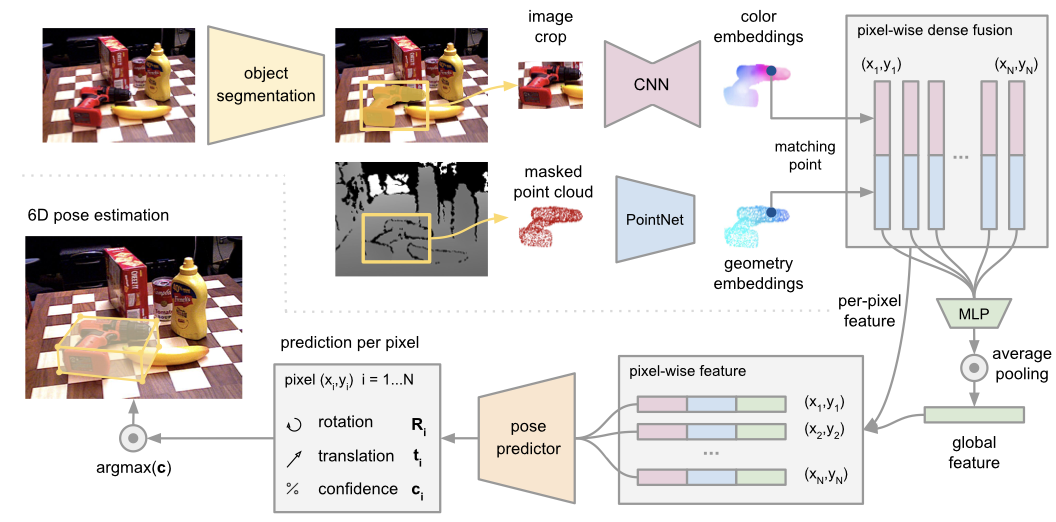}
	\caption{Pipeline of the DenseFusion networks. The network first generates object segmentation masks and 2D bounding boxes from color images. The color-and-depth image pairs are cropped using the bounding boxes and fed into embeddings and fused at each corresponding pixel. The pose predictor estimates a 6D pose for each fused feature and  the predictions are voted to obtain the final 6D object pose. }
	\label{Fig.2}
\end{figure}

Figure \ref{Fig.2} shows the overall architecture of  DenseFusion. The architecture consists of two stages. Firstly,  the target object is detected  in the input color image using  semantic segmentation from \cite{xiang2017posecnn}. After that, the color and depth images are cropped based on the segmentation and  the cropped depth image is transformed to a point cloud using the intrinsic camera matrix. Both cropped images are fed to the second stage.

In the second stage, the cropped color image is fed to a CNN-based network Resnet-18 \cite{he2016deep} encoder followed by $4$ up-sampling layers as decoder to extract color features. The point cloud converted by the cropped depth map is fed into a PointNet-based network \cite{qi2017pointnet} by applying a multi-layer perceptron (MLP)  to produce geometric features. After that, the color and depth features are fused to estimate the 6D object pose based on an unsupervised confidence score. Lastly, the predicted pose is refined by the iterative pose refinement network.

We implement the DenseFusion  network  within the PyTorch framework and the model is trained using Adam optimizer with an initial learning rate at $0.0001$. The iterative pose refinement module contains a $4$ fully connected layers and $2$ refinement iterations is used for the experiments.

\subsection{ASS3D: Adaptive Single-Shot 3D Object Pose Estimation}
\begin{figure}[htb]
	\centering
	\includegraphics[scale=0.27]{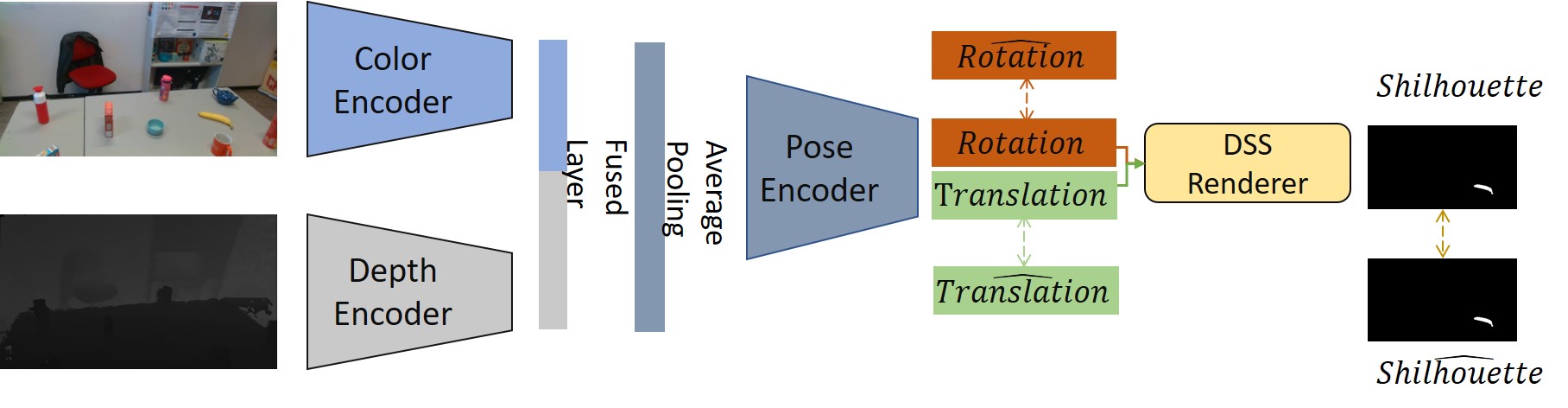}
	\caption{Overall Network Architecture. The color and depth images are processed separately and the extracted features are fused in a later stage. They employ an average-pooling function as the symmetric reduction function. The features are then driven into a pose encoder which eventually directly regresses a rotation and translation. The predicted pose is subsequently used for rendering the object and deriving its projected silhouette. This allows  utilizing an additional supervision signal during training and increase the overall performance of the model.}
	\label{Fig.3}
\end{figure}

Multimodal inputs can improve the performance of various computer vision tasks, but it is usually at the cost of efficiency and increased complexity. In this work, they focus on RGB-D 6D object pose estimation and exploit multimodal inputs using a lightweight fusion scheme which is complemented by multimodal supervision through rendering. In this way, they overcome the complexity of multimodal inputs by transferring it to the model training phase instead of the inference phase. Given the distinct domains that color and depth information resides in, they employ a disentangled architecture, as depicted in Figure \ref{Fig.3}, to process them separately and enable for a learnable fusion scheme.

More specifically, they use two ResNet-34 models as their backbone encoders for extracting features, which are later fused and flattened by an average-pooling function. This approach allows them to associate the geometric feature of each point to its corresponding image feature pixel based on a projection onto the image plane using the known camera intrinsic  parameters as it has been already shown in \cite{wang2019densefusion}. The fused features are then fed into a pose encoder consisting of three fully connected layers that eventually disentangled to 3D rotation and 3D translation heads.   Following the definition of their model’s architecture, they  supervise it using a direct pose regression objective as the weighted sum of two different losses. Particularly, they use a $L2 $  loss $\varepsilon_{t} = ||t- \tilde{t}||$ for the translation and a geodesic distance for the rotation $\varepsilon_{r} = \arccos \frac{trace(R\tilde{R}^{T})-1}{2}$, similar to \cite{Albanis_2020_ECCV_Workshops}. The loss for the predicted pose is then:

\begin{equation}
\varepsilon_{pose} = \lambda_{six_{d}} \varepsilon_{t} +(1- \lambda_{six_{d}}) \varepsilon_{r},  
\end{equation}

\noindent where the weight $\lambda_{six_{d}}$ acts as a regularization term. This is complemented by a silhouette loss which is enabled by a point splatting differentiable renderer \cite{yifan2019differentiable}. They transform the 3D vertices $ \nu\in \mathbb{R}^3$ of each object's point cloud using the predicted pose. The differentiable point cloud renderer then renders the transformed model's silhouette, which is used along with the ground truth annotated silhouettes as an additional supervision signal. Instead of using a traditional intersection over union (IoU)  loss, they apply a Gaussian smooth silhouette loss as defined in  \cite{Albanis_2020_ECCV_Workshops} for their silhouette loss:
\begin{equation}	
\varepsilon_{silhouette}  = \frac{1}{N} \sum_{\in \Omega } S \astrosun    \mathscr{S} (\tilde{S}) + \tilde{S} \astrosun \mathscr{S}(S),
\end{equation}

\noindent where $S$ is a Gaussian smoothing function. The silhouette loss is a smoother objective function compared to the common IoU loss, while it takes into account the ground truth silhouette simultaneously, offering that way a fully symmetric objective. However, the most appropriate Gaussian filter to be used is dependent on each object shape, and can also vary during training, offering higher precision as the model converges. Towards that end, they use a new adaptive filter by also learning the standard deviation of the Gaussian during training. Their final learning objective is a weighted sum of the aforementioned losses: 
\begin{equation}
\varepsilon_{total} = \lambda_{pose}\varepsilon_{pose} +\lambda_{silhouette}\varepsilon_{silhouette}
\label{giorgos}
\end{equation}

It is apparent that the introduction of the weights $\lambda_{pose}$ and
$\lambda_{silhouette}$ will introduce similar challenges as aforementioned (i.e. finding the best combination for each object will be challenging and time-consuming). Motivated by that, they treat those two weights as learnable parameters adding them to the learning objective. Thus, the weights are able to adapt to the various objects and, additionally, to better regularize the two losses during training.

Finally, the model is trained for 100 epochs on a GeForce RTX 2080 TI 11 GB. All the images (i.e. color and depth) are resized to $320 \times 180$ resolution, and the batch size is set to $16$. For optimizing the model’s parameters they use the Adam optimizer with a learning rate of $1 \times 10^{-4}$. Additionally, learnable Gaussian standard deviation and the weights of \ref{giorgos} are optimized with a SGD optimizer with a learning rate of $1 \times 10^{-5}$.

\subsection{GraphFusion: 6D object pose estimation with graph based multi-feature fusion }
They propose a  graph based  multi-feature fusion network  to improve   6D pose prediction performance, which combines  effective  feature extraction networks  and  a  graph attention network (GAT) \cite{velivckovic2017graph} to fully exploit the relationship between visual and geometric features.
\setlength{\textfloatsep}{10pt plus 1.0pt minus 2.0pt}
\begin{figure}[htb]
	\centering
	\includegraphics[width=0.48\textwidth]{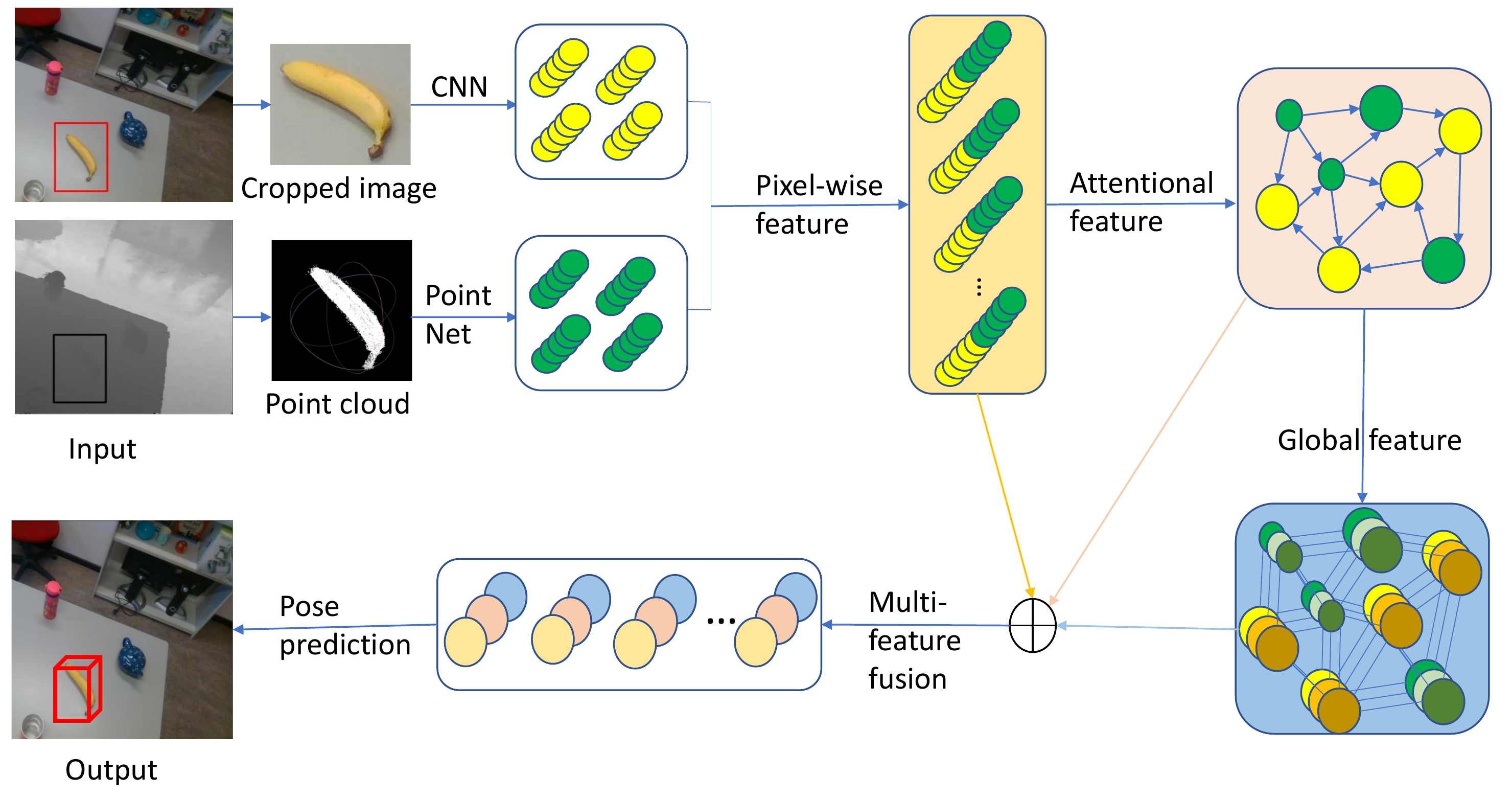}
	\caption{Overview of the graph based pose estimation architecture. The input of their networks are captured color-and-depth images pairs. These images are cropped with the semantic segmentation architecture. After that,  the visual and geometric features are extracted and fused by  a graph attention network which is introduced to exploit the fusion strategy between color and geometric features.  The 6D object pose and its corresponding confidence score are predicted by the fused features and the final pose is  chosen  based on the confidences. }
	\label{Fig.5}
\end{figure}

The aim of their approach is to achieve the real-time 6D pose estimation, using RGB-D images as input, as shown in Figure\ref{Fig.5}. Handcrafted features such as SIFT or ORB are key factors for classical methods to estimate 6D poses. However, it is difficult to estimate 6D poses for texture-less objects. Instead of relying on improving handcrafted features, they learn more robust features and semantic cues  by applying deep learning models.

They  use a Convolutional Neural Network (CNN) based encoder-decoder architecture to learn visual features from color images. To extract geometric features from the depth map, they  first convert the depth map to the point cloud using the camera intrinsic matrix. There are two ways to process the point cloud.   Classic approaches  often convert point cloud data into regular grids by projecting  3D data into 2D images or splitting raw data into 3D voxel grids. Then they process the transformed data using approaches  based  on regular data. Other approaches are to directly process each point in the  point cloud.  PointNet \cite{qi2017pointnet} is the first one to apply this idea,  which achieves permutation invariance by use of  a  symmetric function. Instead of transforming to regular data, they use PointNet-based network to extract geometric features from the point cloud.

Even with learned features that contains the visual appearance and geometry structure information, accurate 6D object pose also depends on the fused  features.    To effectively fuse  features, they introduce a graph attention  based  framework  to exploit relationship between visual and geometric features, as opposed to prior works which just concatenates these features.  Combining the insights above, their approach works as follows:

The input are captured color-and-depth image pairs and a  semantic segmentation architecture from \cite{xiang2017posecnn} is used to segment the target object and  crop the color and depth images. Next, the visual  features are extracted by a CNN-based network and geometric representations are  computed from the  point cloud  using PointNet.  The point cloud is generated by converting its corresponding depth map. With these features, a graph attention network is introduced to perform the fusion  between color and geometric features. After that, the 6D object pose and its corresponding confidence score are predicted by the fused features, one pose per fused feature. Then, the pose with the highest confidence is chosen as the estimated pose. Lastly, the 6D pose  is further improved by iterative pose refinement. 

\section{Results}

\subsection{Overall performance}
The overall performance  of DenseFusion, ASS3D, GraphFusion without refinement (GraphFusion\_wo) and GraphFusion  is shown in Table \ref{table.1} and Table \ref{table.2}. DenseFusion and ASS3D are proposed from two different research groups, and GraphFusion without refinement (GraphFusion\_wo) and GraphFusion is proposed from one research group.  We use ADD, ADD-S and the area under ADD curve (AUC)   to measure the prediction.

\begin{table}[htp]
	
	\centering
	\caption{Quantitative evaluation of the 6D pose (ADD and ADD-S).}
	\begin{adjustbox}{ width=0.48\textwidth}
		\begin{tabular}{l|c|c|c|c|c|c|c|c}
			\hline
			\multirow{2}{*}{} & \multicolumn{2}{c|}{DenseFusion} & \multicolumn{2}{c|}{   ASS3D}             & \multicolumn{2}{c|}{GraphFusion\_wo}                         & \multicolumn{2}{c}{GraphFusion}              \\ \cline{2-9} 
			& ADD& ADD-S& ADD & ADD-S & ADD & ADD-S &ADD &ADD-S \\ \hline
			banana   & \textbf{0.86}   & 0.86    & 0.70    & 0.75    & 0.76   &  0.80   &  0.83    &  \textbf{0.87}   \\ \hline
			biscuit\_box      &  0.91   & 0.95    &  0.78   &  0.88   & 0.80   &  0.84   &  \textbf{0.93}     &  \textbf{0.96}   \\ \hline
			chips\_can        &  0.56   & 0.94    &  \textbf{0.75}   & 0.85    &  0.53   &  0.57   &  0.69     & \textbf{0.97}    \\ \hline
			cookie\_box       &  \textbf{0.62}   & 0.74    & 0.49    & 0.66    & 0.51    & 0.56    &  0.61     &  \textbf{0.75}   \\ \hline
			gingerbread\_box  &  0.87   & 0.94    & 0.63     & 0.86   &  0.79   &  0.83   &  \textbf{0.90}    &  \textbf{0.95}   \\ \hline
			milk\_box         &  0.50   & \textbf{0.81}    &  0.58   & 0.62    & 0.52    &  0.53  &  \textbf{0.66}     &  0.77   \\ \hline
			pasta\_box        &  0.77   & 0.91    &  0.63   & 0.72    & 0.76    &  0.78   &  \textbf{0.84}     &  \textbf{0.96}   \\ \hline
			vacuum\_cup       &  0.61   & 0.90    & \textbf{ 0.65}   &  0.75   &  0.51   & 0.53    &  0.63     &   \textbf{0.97}  \\ \hline
			MEAN              &  0.71   & 0.88    &  0.65    & 0.76    & 0.65    &  0.68   &  \textbf{0.76}     &  \textbf{0.90 }  \\ \hline
		\end{tabular}
	\end{adjustbox}
	\label{table.1}	
\end{table}

\begin{table}[htb]
	\centering
	\caption{The 6D pose estimation accuracy in terms of the area under AUC.}
	\begin{adjustbox}{ width=0.48\textwidth}
		\begin{tabular}{l|c|c|c|c}
			\hline
			& DenseFusin & ASS3D & GraphFusion\_wo & GraphFusion \\ \hline
			banana            &  \textbf{0.77}   &  0.66   &  0.66   & 0.75         \\ \hline
			biscuit\_box      &  0.77   &  0.74   &  0.68   &  \textbf{0.79}         \\ \hline
			chips\_can        & 0.74    &  0.72   &  0.55   & \textbf{0.76 }        \\ \hline
			cookie\_box       &  \textbf{0.67}   &  0.56   &  0.53   &  0.66        \\ \hline
			gingerbread\_box  &  0.76   &  0.71   &  0.64   &  \textbf{0.78}        \\ \hline
			milk\_box         &  0.66   & 0.51    &  0.52   & \textbf{0.67}         \\ \hline
			pasta\_box        &  0.74   &  0.61   &  0.61   &  \textbf{0.77}        \\ \hline
			vacuum\_cup       &  0.71   &   0.64  &  0.63   &  \textbf{0.74}        \\ \hline
			MEAN              &  0.72   &  0.64   &  0.60   &  \textbf{0.74}        \\ \hline
		\end{tabular}
	\end{adjustbox}
	\label{table.2}
\end{table}

We can see that GraphFusion  achieves the best performance.  GraphFusion outperforms ASS3D and DenseFusion $11\%$ and $5\%$ in terms of ADD, respectively. Besides, evaluation results show that  DenseFusion, ASS3D and GraphFusion allow transforming  models trained in synthesized  datasets  to  real captured datasets without  domain adaptation.

\begin{figure}[htb]
	\centering  
	
	\subfigure[banana]{                    %
		\begin{minipage}{.23\textwidth}
			\centering                                                          %
			\includegraphics[scale=0.3]{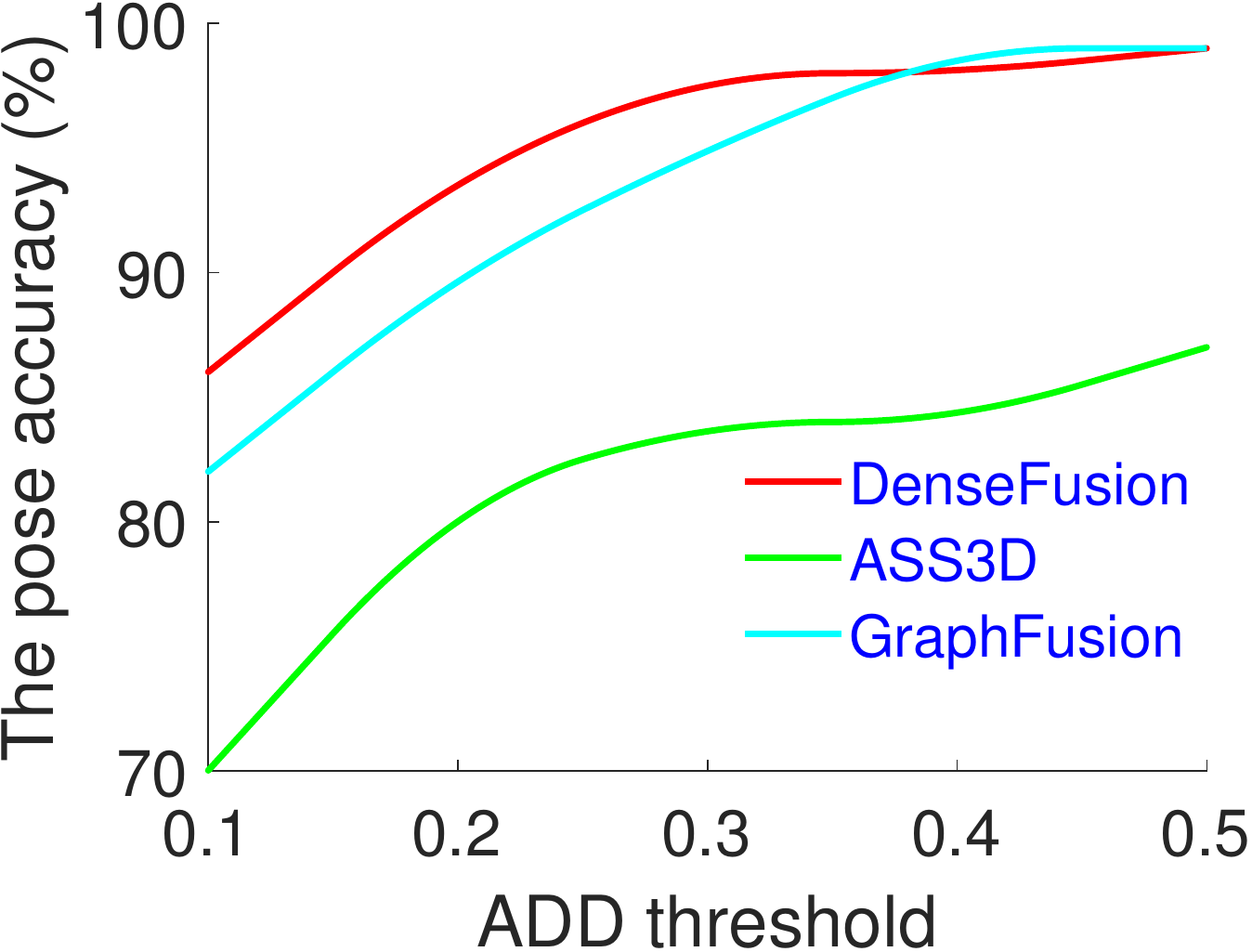}               %
	\end{minipage}}	
	\subfigure[biscuit\_box]{      
		\begin{minipage}{.23\textwidth}
			\centering                                                          %
			\includegraphics[scale=0.3]{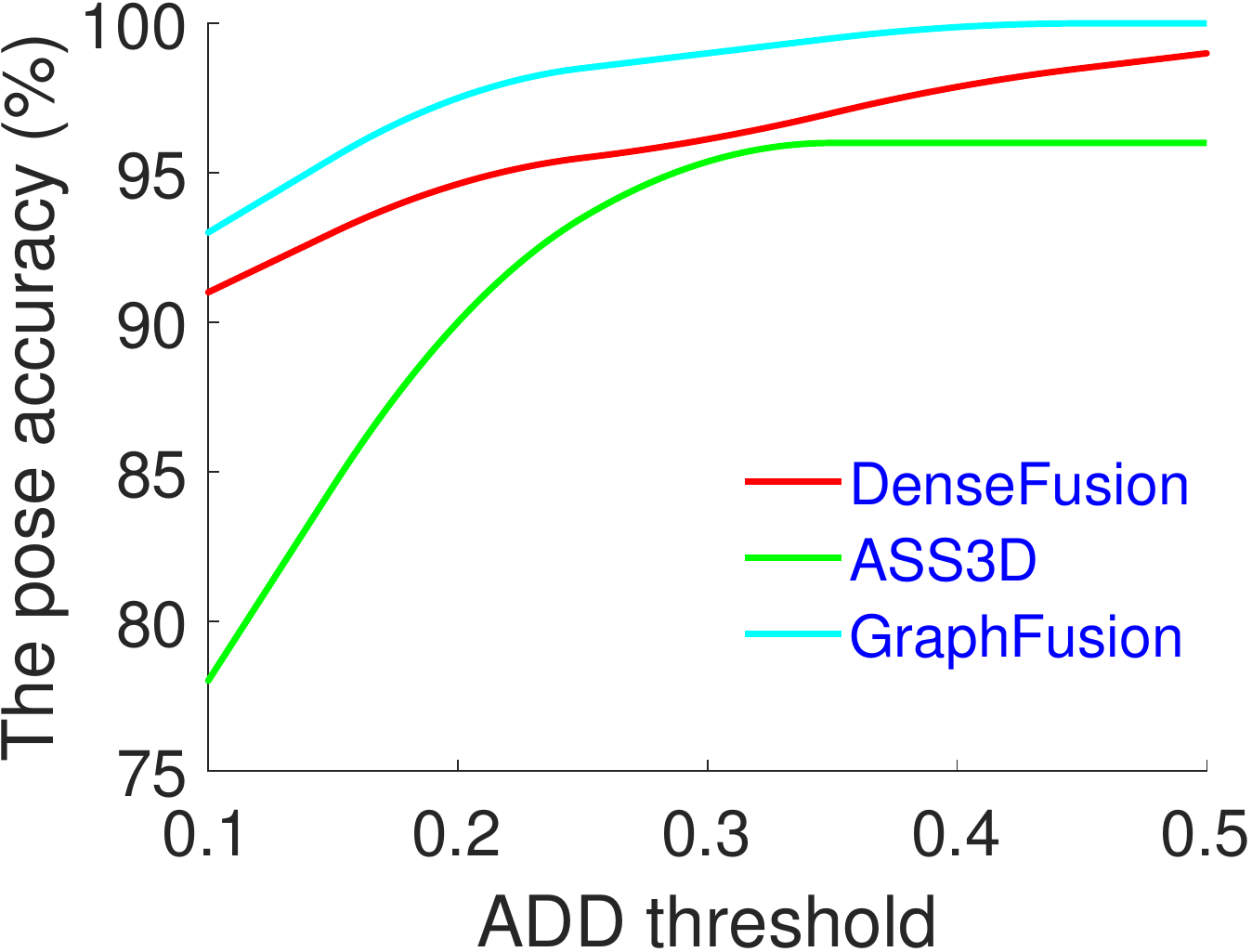}                %
	\end{minipage}}
	\subfigure[chips\_can]{      
		\begin{minipage}{.23\textwidth}
			\centering                                                          %
			\includegraphics[scale=0.3]{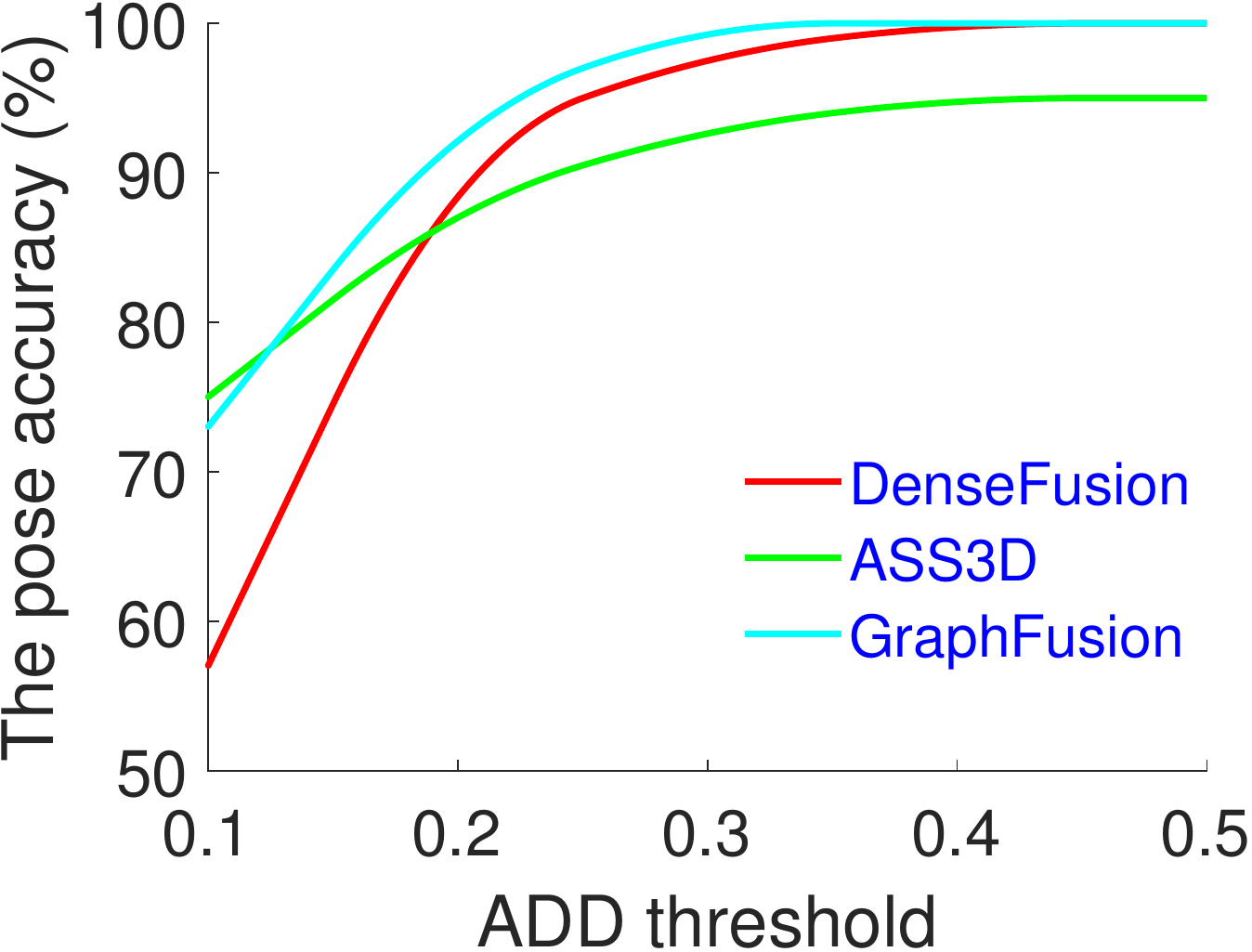}                %
	\end{minipage}}
	\subfigure[cookie\_box]{     
		\begin{minipage}{.23\textwidth}
			\centering                                                          %
			\includegraphics[scale=0.3]{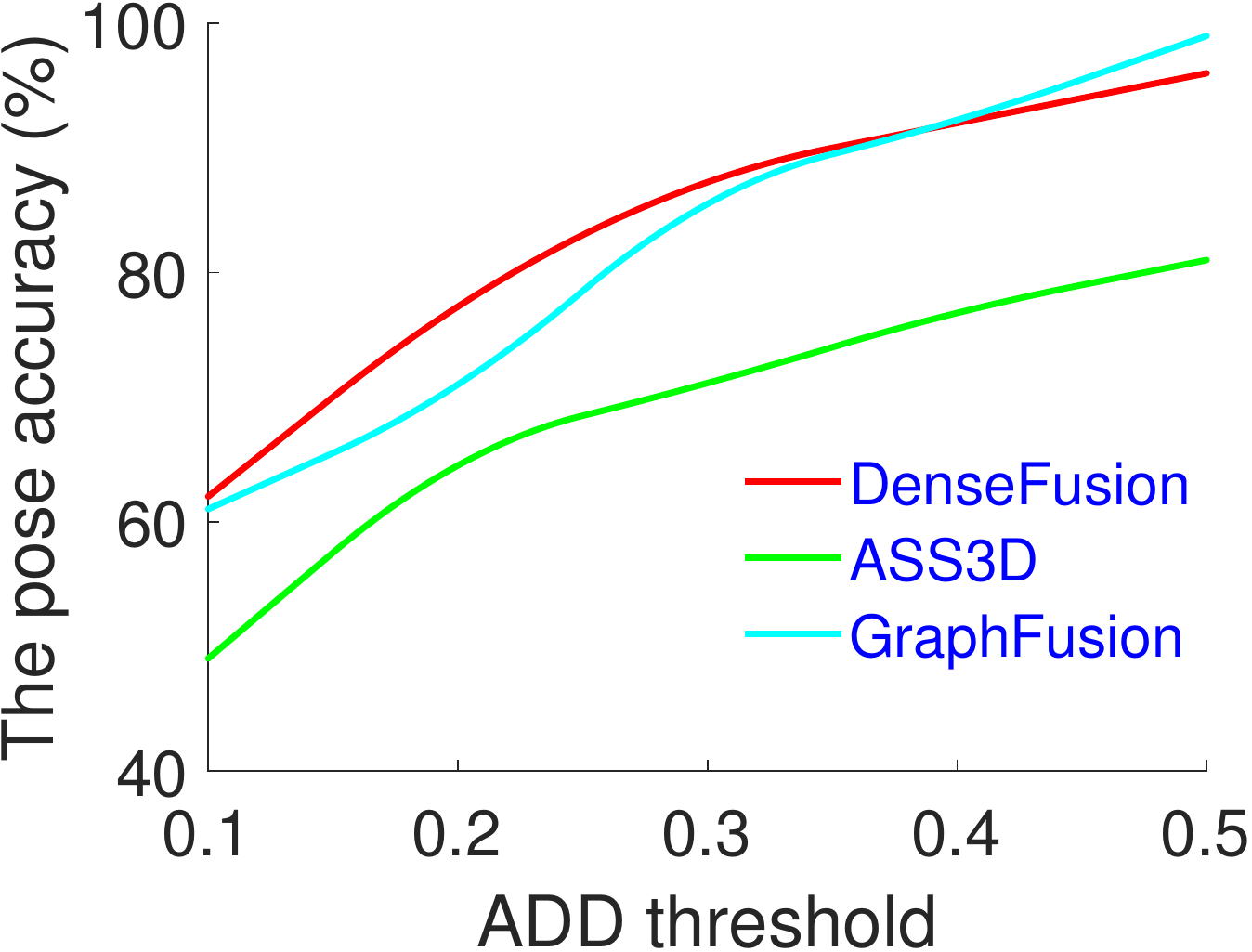}                %
	\end{minipage}}	
	\subfigure[gingerbread\_box]{     
		\begin{minipage}{.23\textwidth}
			\centering                                                          %
			\includegraphics[scale=0.3]{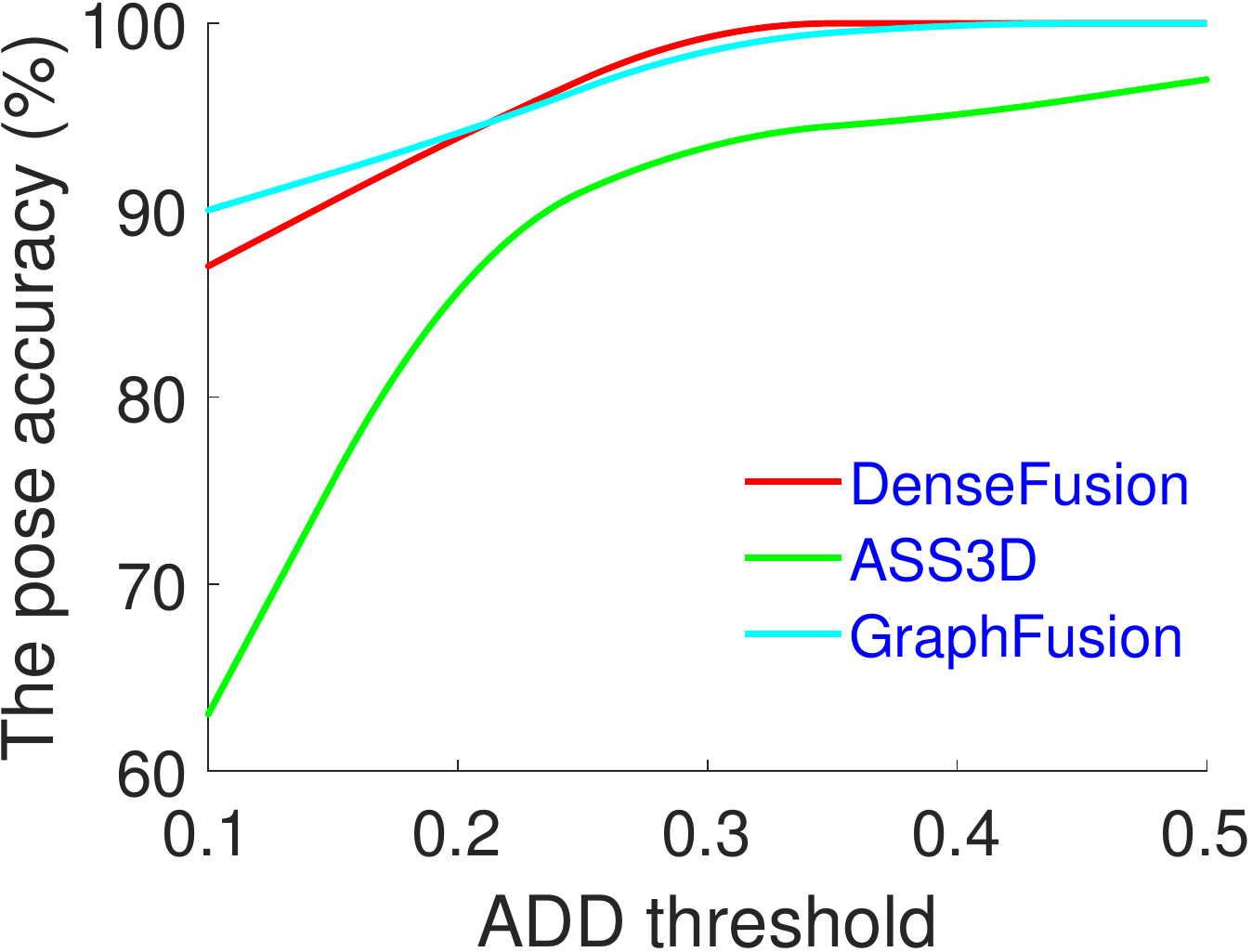}                %
	\end{minipage}}
	\subfigure[milk\_box]{     
		\begin{minipage}{.23\textwidth}
			\centering                                                          %
			\includegraphics[scale=0.3]{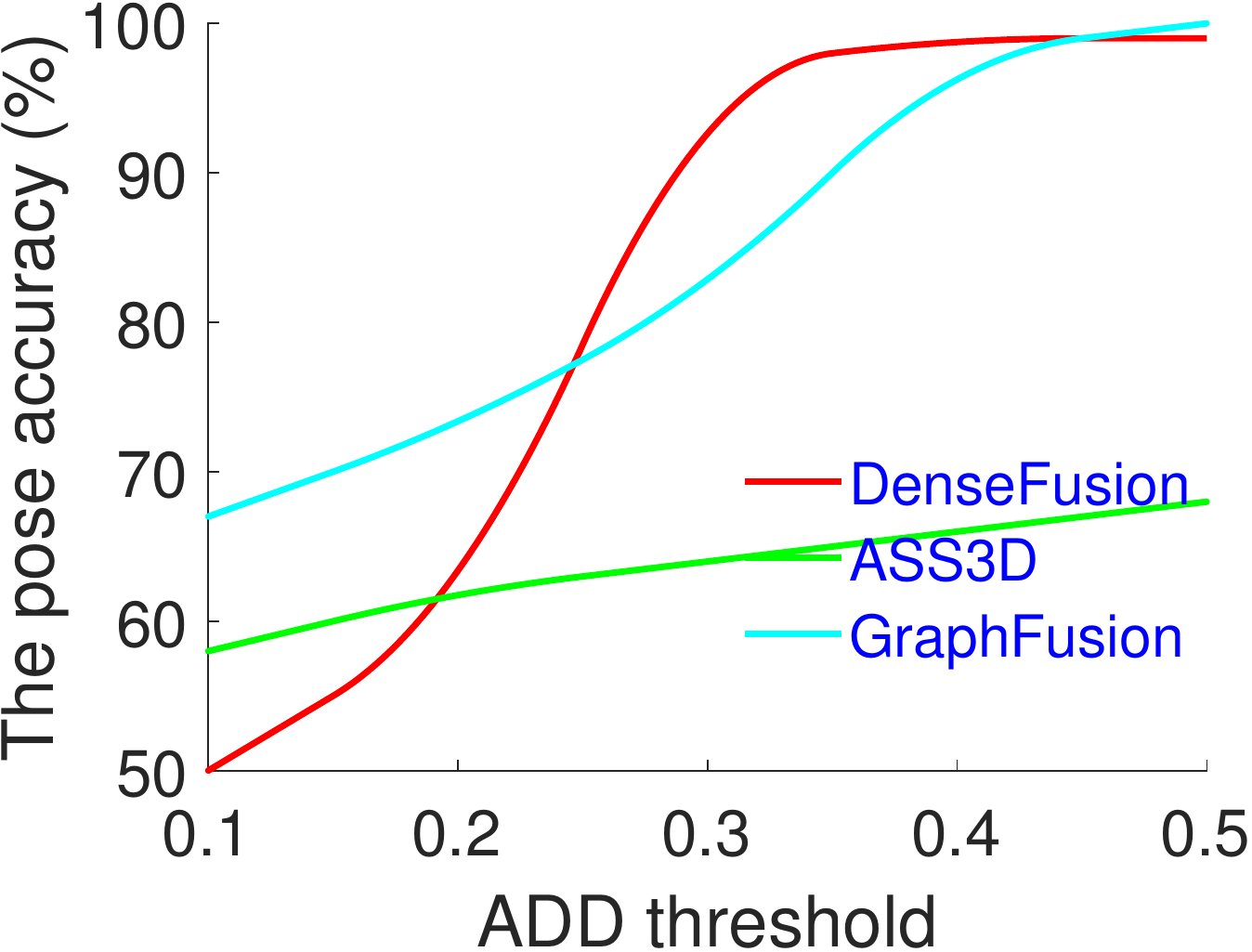}                %
	\end{minipage}}
	\subfigure[pasta\_box]{     
		\begin{minipage}{.23\textwidth}
			\centering                                                          %
			\includegraphics[scale=0.3]{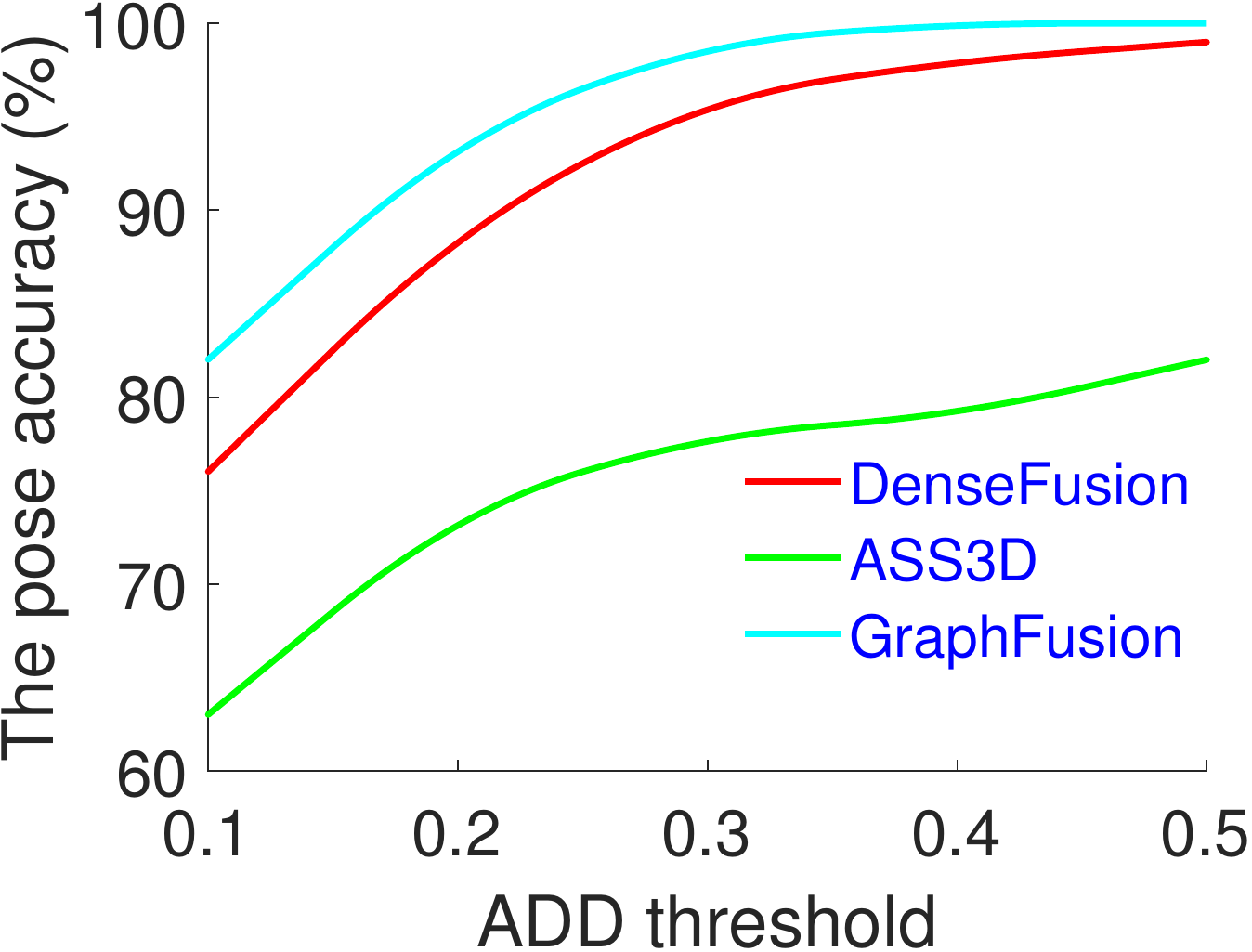}                %
	\end{minipage}}
	\subfigure[vacuum\_cup]{     
		\begin{minipage}{.23\textwidth}
			\centering                                                          %
			\includegraphics[scale=0.3]{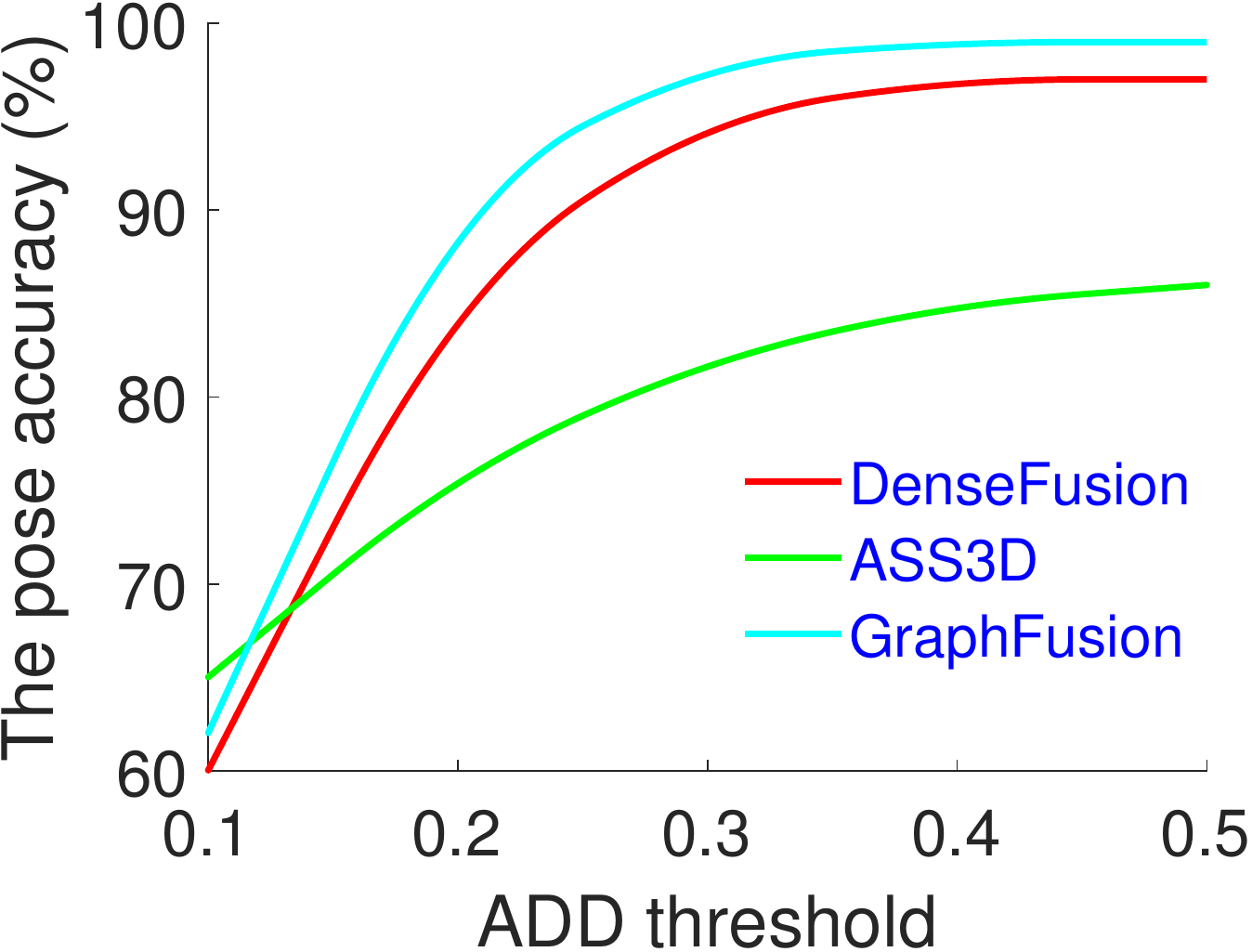}                %
	\end{minipage}}
	\caption{The successful  rate of pose estimation in terms of ADD.}                                           
	\label{Fig.6}                                                        %
\end{figure}

\subsection{Ablation studies}

We present ablation studies to help better understand the functionalities of different network architectures.

\textbf{Effectiveness of pose refinement.}  From Table \ref{table.1} and Table \ref{table.2} we can see that compared with ASS3D and GraphFusion without pose refinement, DenseFusion and GraphFusion that perform iterative pose refinement are able to further improve the accuracy of the 6D pose.  The effectiveness is further verified by Figure \ref{Fig.6}.  Figure \ref{Fig.6} shows the  successful pose rate  measured by ADD for $8$ objects, which is obtained by varying the ADD threshold.  DenseFusion and GraphFusion    outperform other approaches by a large margin, especially when the threshold is small.  

\begin{figure}[htb]
	\centering  	
	\subfigure[banana]{                    %
		\begin{minipage}{.23\textwidth}
			\centering                                                          %
			\includegraphics[scale=0.28]{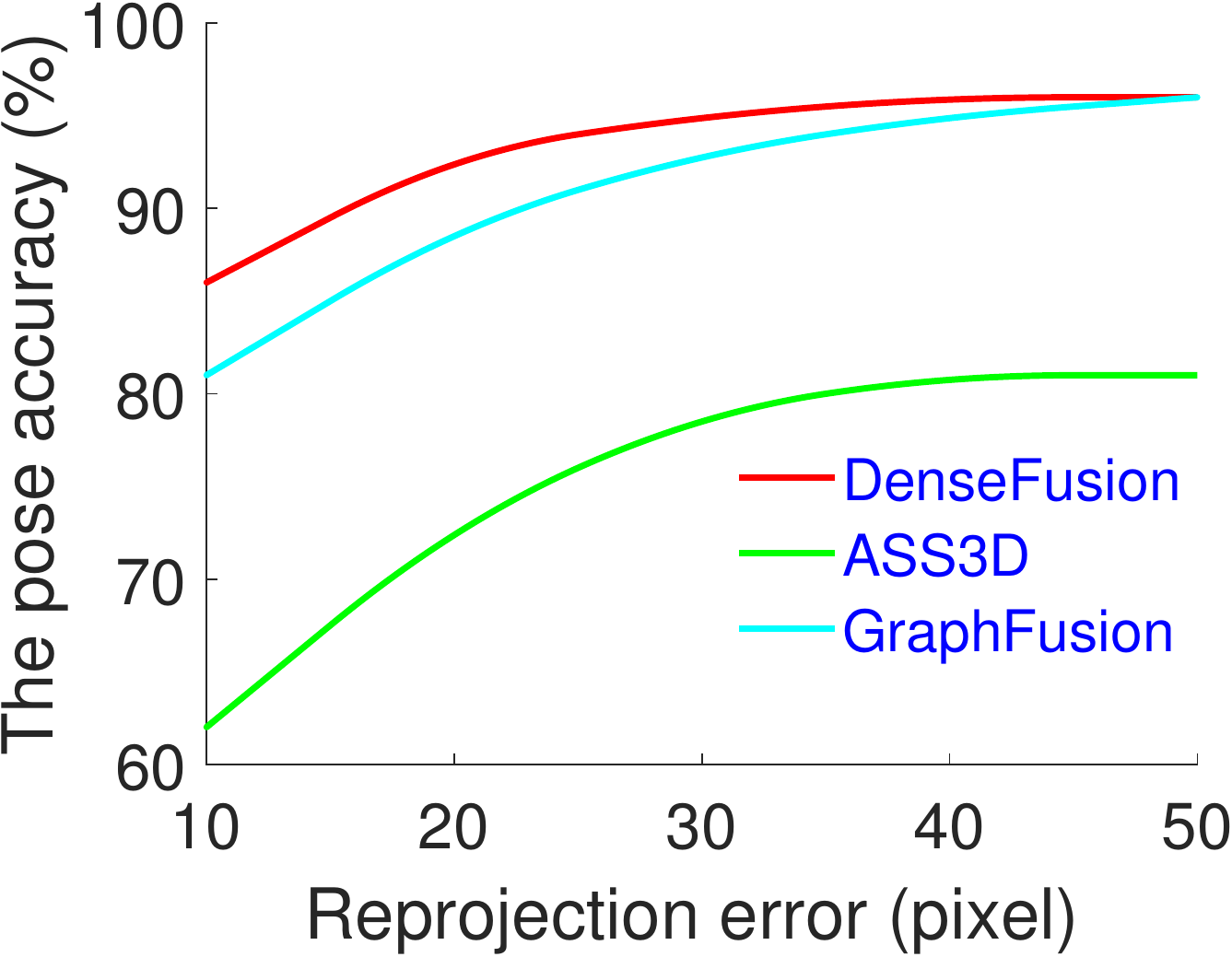}               %
	\end{minipage}}	
	\subfigure[biscuit\_box]{      
		\begin{minipage}{.23\textwidth}
			\centering                                                          %
			\includegraphics[scale=0.3]{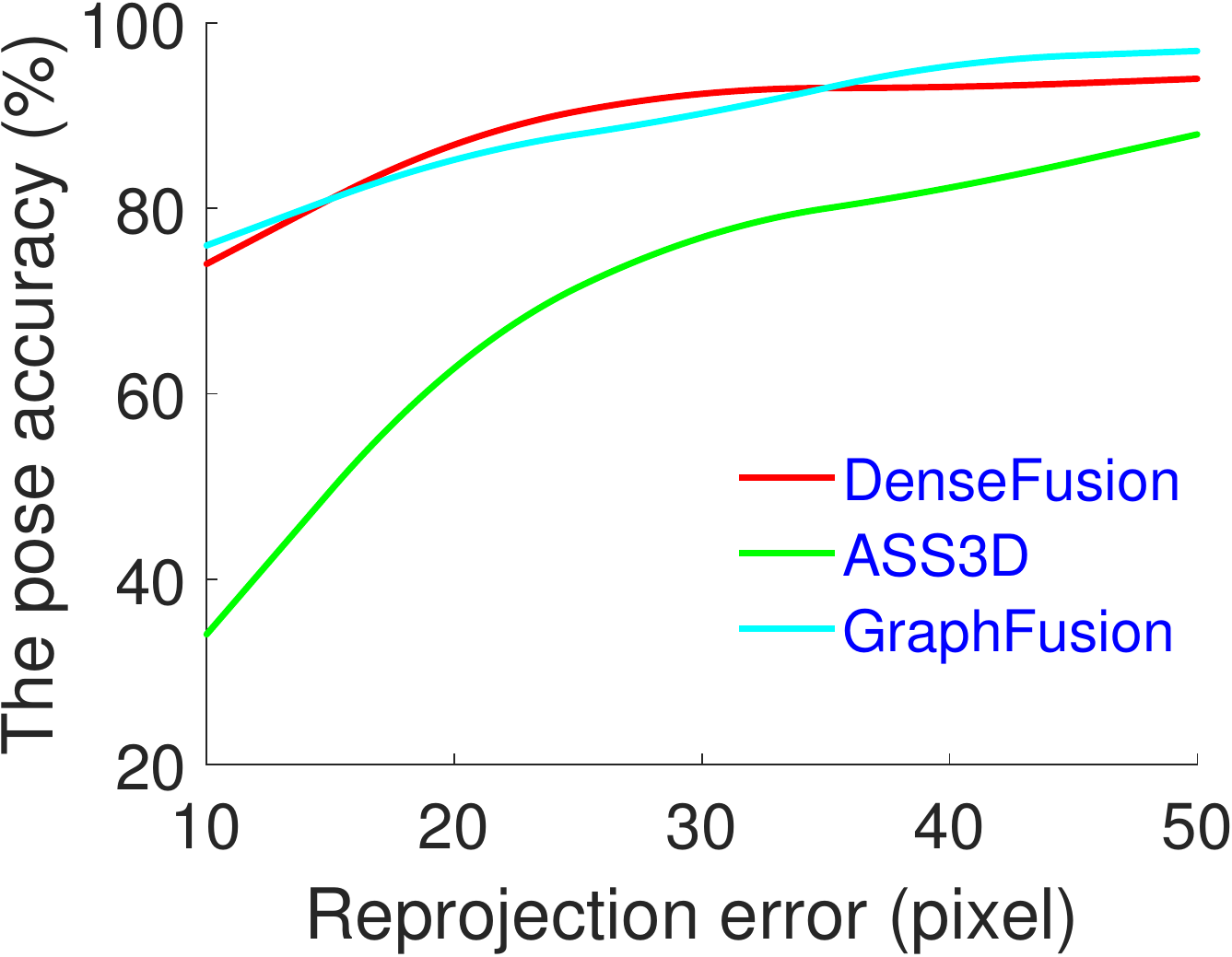}                %
	\end{minipage}}
	\subfigure[chips\_can]{      
		\begin{minipage}{.23\textwidth}
			\centering                                                          %
			\includegraphics[scale=0.3]{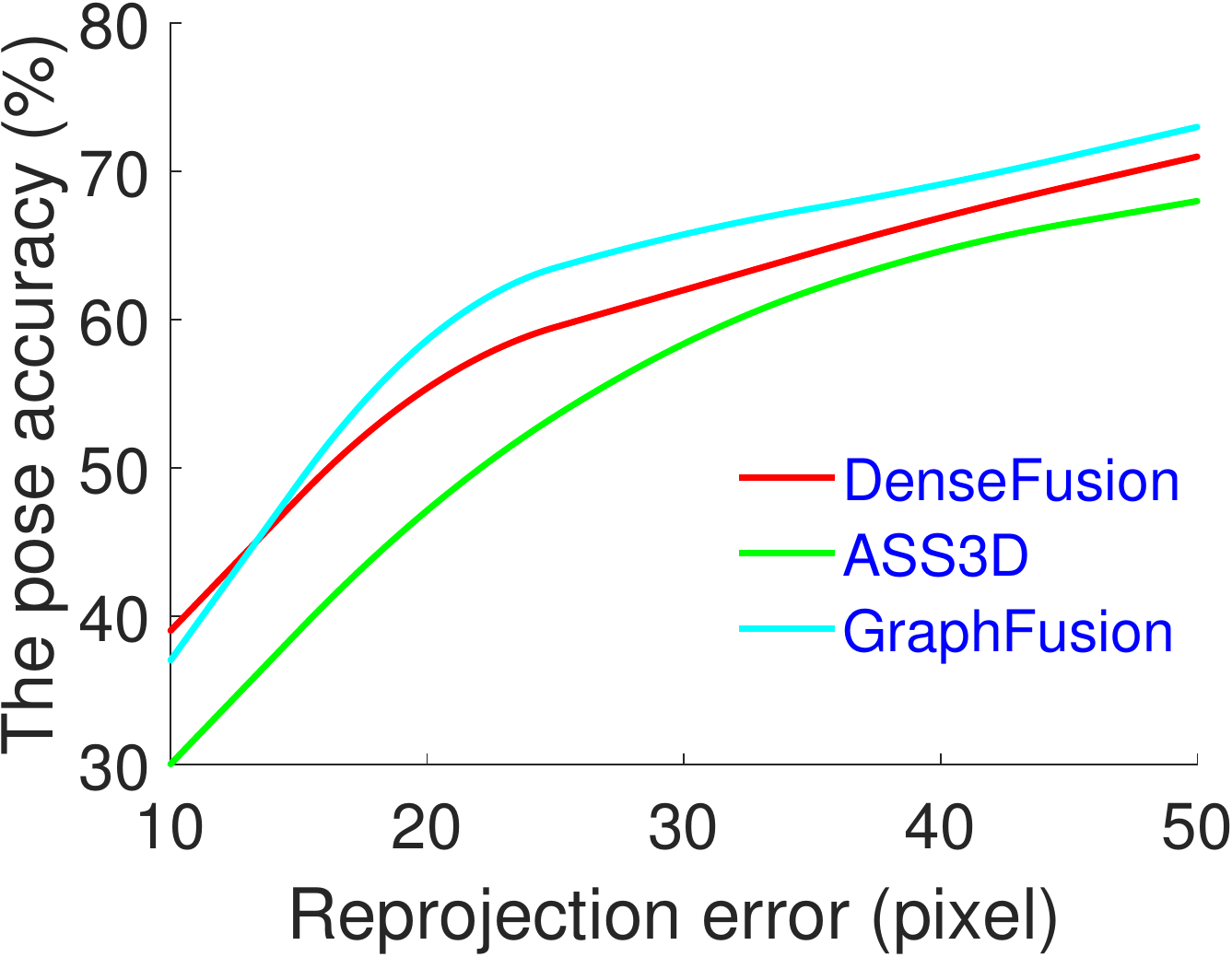}                %
	\end{minipage}}
	\subfigure[cookie\_box]{     
		\begin{minipage}{.23\textwidth}
			\centering                                                          %
			\includegraphics[scale=0.3]{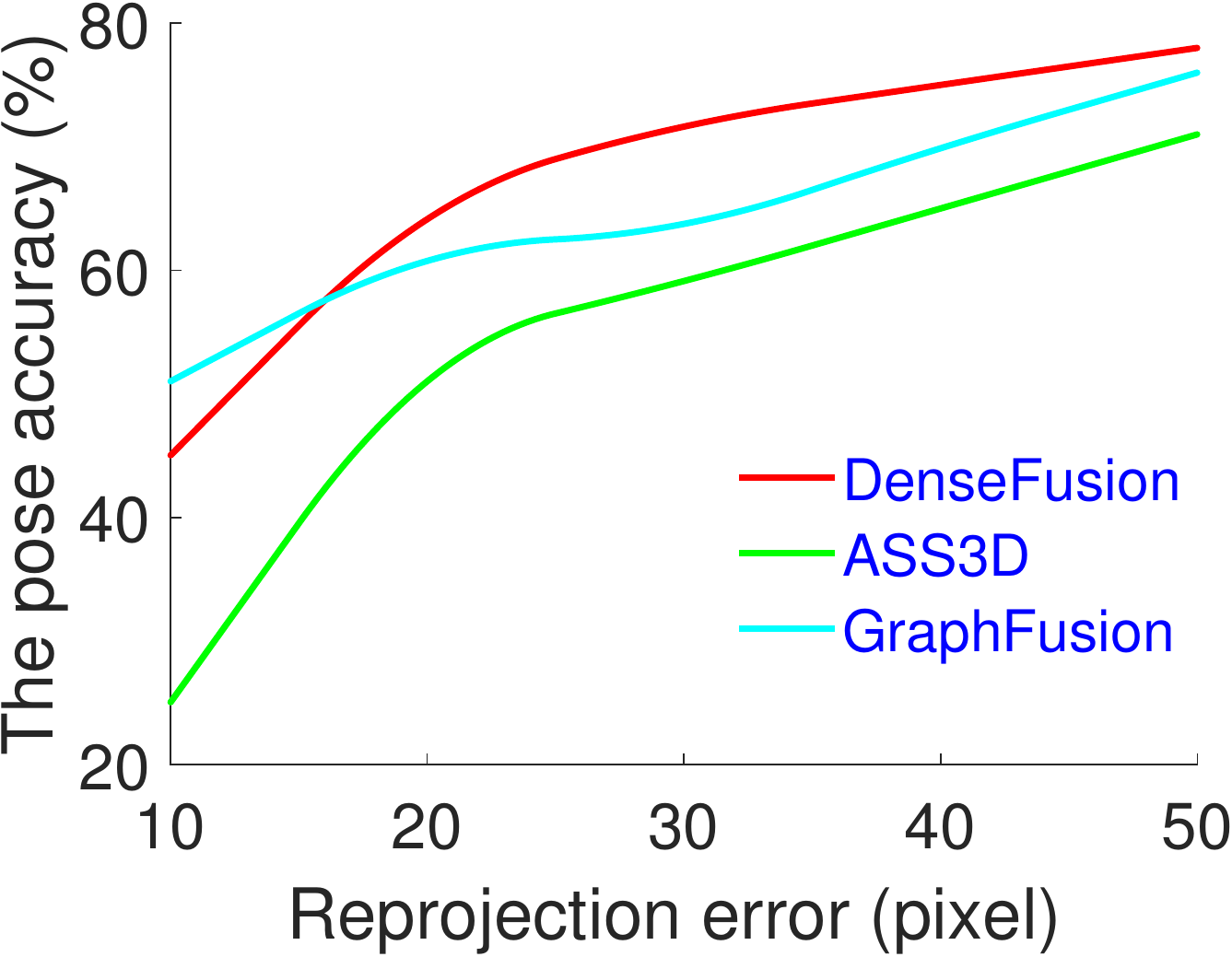}                %
	\end{minipage}}	
	\subfigure[gingerbread\_box]{     
		\begin{minipage}{.23\textwidth}
			\centering                                                          %
			\includegraphics[scale=0.3]{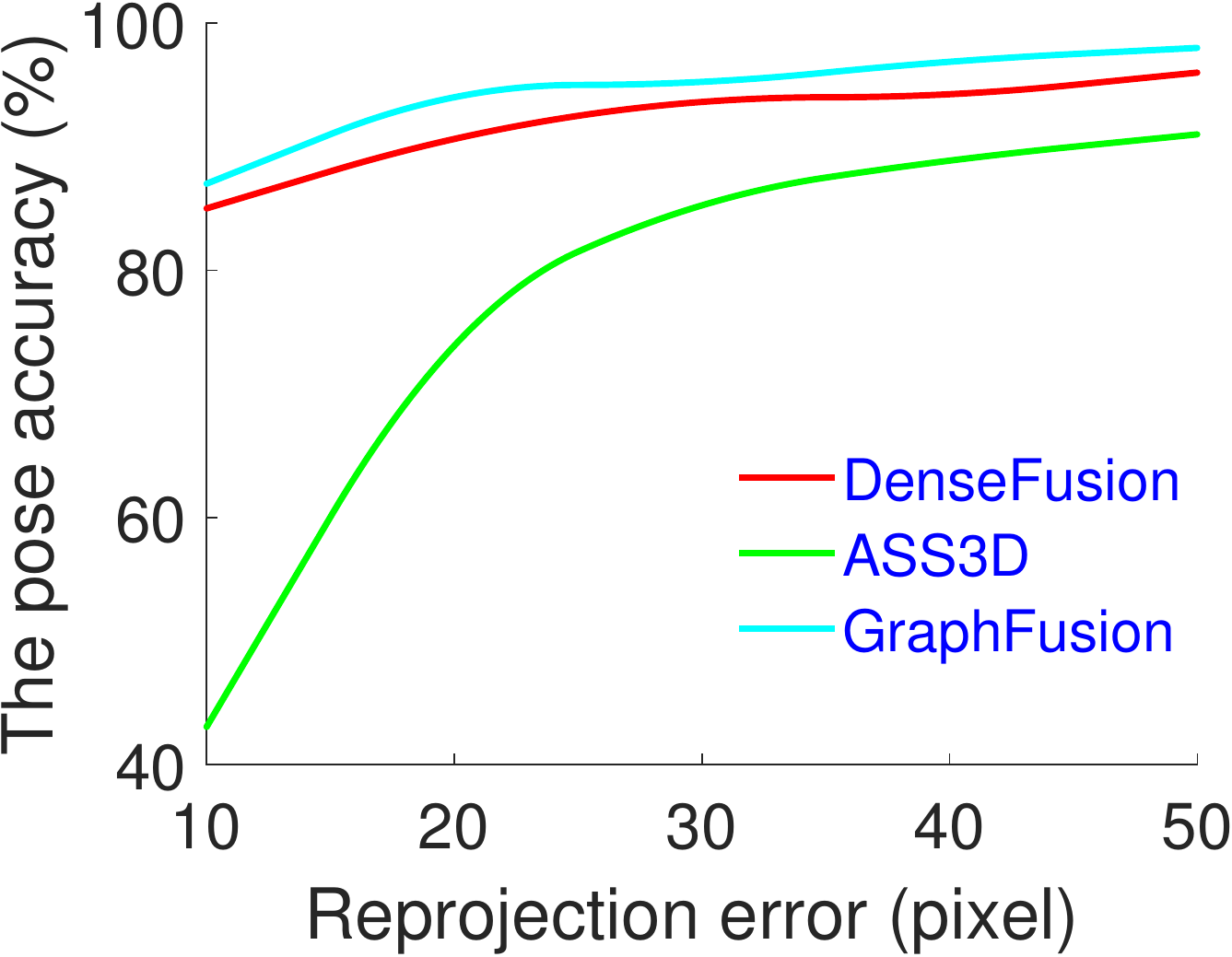}                %
	\end{minipage}}
	\subfigure[milk\_box]{     
		\begin{minipage}{.23\textwidth}
			\centering                                                          %
			\includegraphics[scale=0.3]{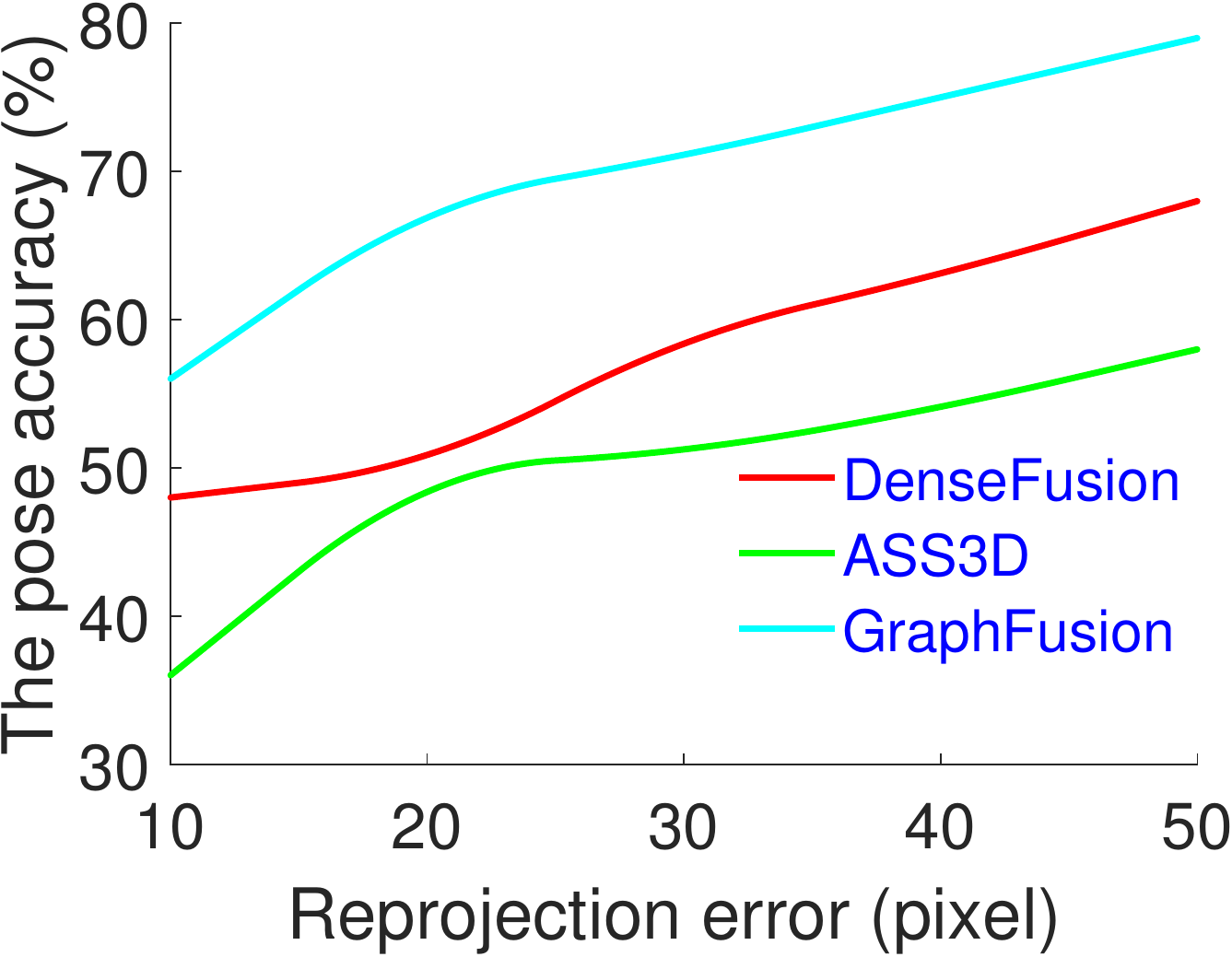}                %
	\end{minipage}}
	\subfigure[pasta\_box]{     
		\begin{minipage}{.23\textwidth}
			\centering                                                          %
			\includegraphics[scale=0.3]{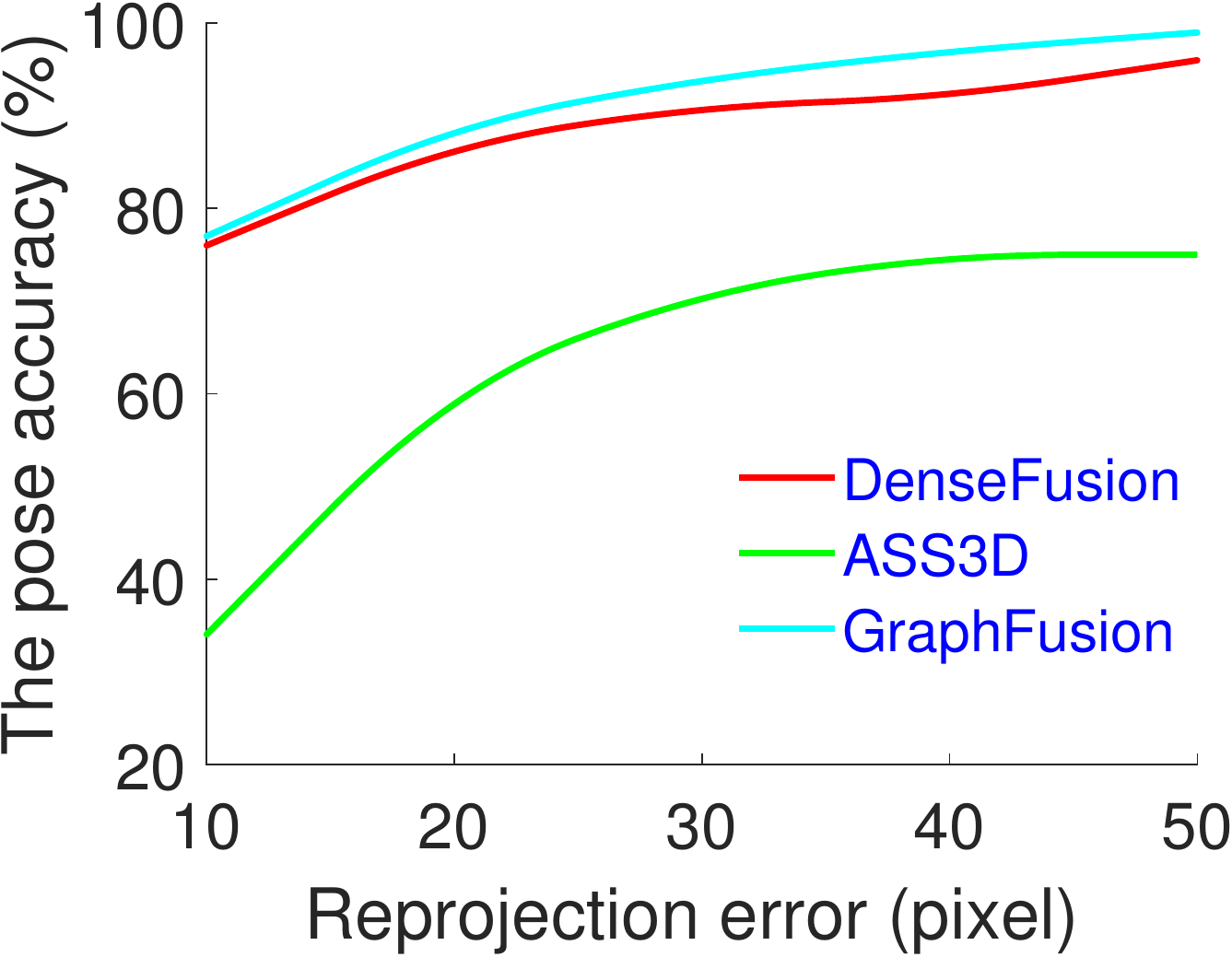}                %
	\end{minipage}}
	\subfigure[vacuum\_cup]{     
		\begin{minipage}{.23\textwidth}
			\centering                                                          %
			\includegraphics[scale=0.3]{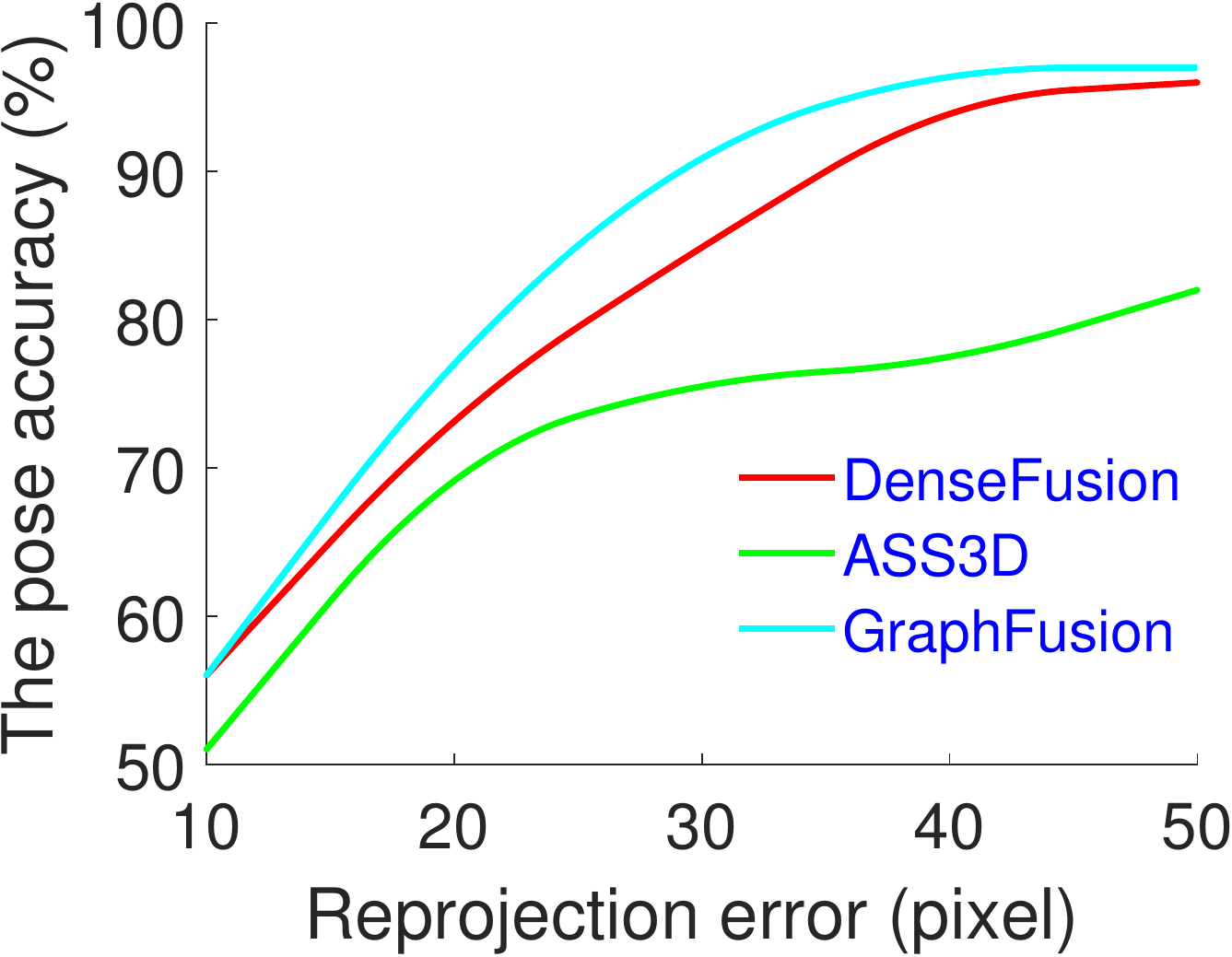}                %
	\end{minipage}}
	\caption{The successful  rate of pose estimation in terms of reprojection errors.}
	\label{Fig.7}                                                        %
\end{figure}

\textbf{Effectiveness of multi-feature fusion.} Apart from the  successful pose rate generated by ADD, we calculate the successful pose rate by varying the reprojection error threshold, as shown in  Figure \ref{Fig.7}. From Figure \ref{Fig.6} and Figure \ref{Fig.7} we can see that  GraphFusion  is superior to  other approaches, which indicates that fusion mechanism considering the relationship between color and geometric features has a clear advantage over methods  ignoring the correlation information between RGB-D images.  As   Figure \ref{Fig.7} makes clear that the performance of   ASS3D degrade significantly as the reprojection error decreases. In contrast, the performance of  DenseFusion and GraphFusion has a smaller decrease. 

\textbf{Time efficiency and accuracy robustness of the single shot model.}  Compared with DenseFusion and GraphFusion, ASS3D estimates the 6D pose in a single, consecutive network pass. It runs faster than other approaches, as show  in Table \ref{table.3} which compares the time efficiency among different methods. In particular, ASS3D runs at least 4 times faster than GraphFusion. Besides, for the texture-less objects such as milk box, ASS3D is more robust and has a better performance as shown in Figure \ref{Fig.6}.  

\begin{table}[h]
	\centering
	\caption{Comparison of the computational run time among different approaches (second per frame).}
	\begin{adjustbox}{ width=0.45\textwidth}
		\begin{tabular}{l|c|c|c}
			\hline
			& DenseFusin & ASS3D  & GraphFusion \\ \hline
			banana            &  0.03   & \textbf{0.01}   &0.04       \\ \hline
			biscuit\_box      &  0.03   & \textbf{0.01}   &  0.04        \\ \hline
			chips\_can        &  0.03   & \textbf{0.01}   & 0.04         \\ \hline
			cookie\_box       &  0.03   & \textbf{0.01}   & 0.04         \\ \hline
			gingerbread\_box  &  0.03   &\textbf{0.01}    & 0.04         \\ \hline
			milk\_box         &  0.03   & \textbf{0.01}  & 0.04         \\ \hline
			pasta\_box        &  0.03   & \textbf{0.01}  & 0.04         \\ \hline
			vacuum\_cup       &  0.03   & \textbf{0.01} & 0.04         \\ \hline
			MEAN              &  0.03   & \textbf{0.01}  & 0.04         \\ \hline
		\end{tabular}
	\end{adjustbox}
	\label{table.3}
\end{table}

\begin{figure*}[htb]
	\centering  
	
	\subfigure[DenseFusion]{                    %
		\begin{minipage}{.9\textwidth}
			\centering                                                          %
			\includegraphics[scale=0.5]{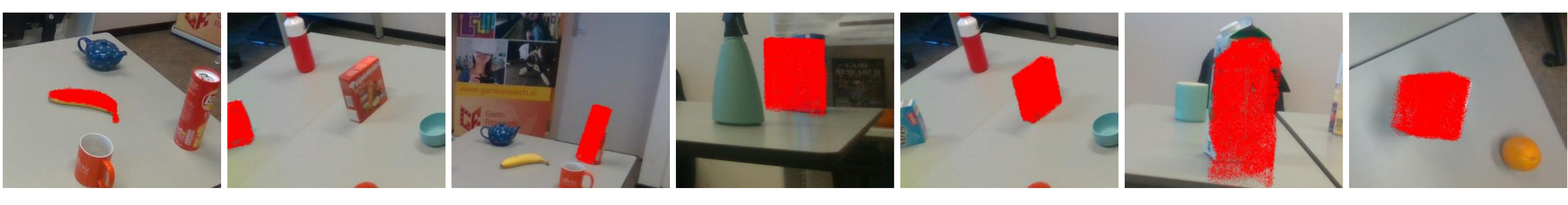}               %
	\end{minipage}}	
	\subfigure[ASS3D]{      
		\begin{minipage}{.9\textwidth}
			\centering                                                          %
			\includegraphics[scale=0.5]{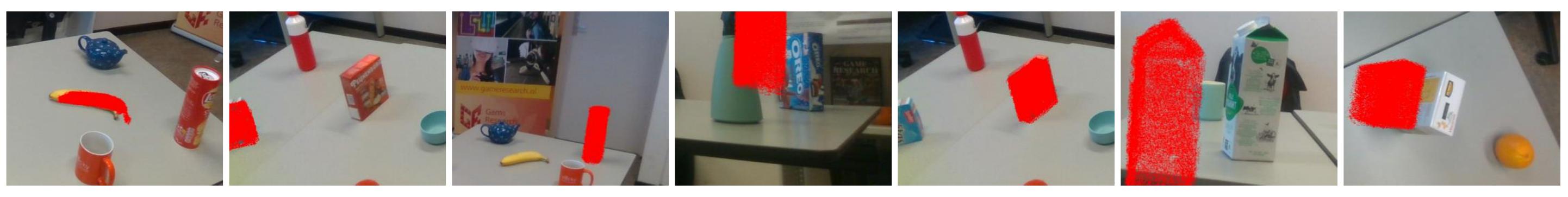}                %
	\end{minipage}}
	\subfigure[GraphFusion]{     
		\begin{minipage}{.9\textwidth}
			\centering                                                          %
			\includegraphics[scale=0.5]{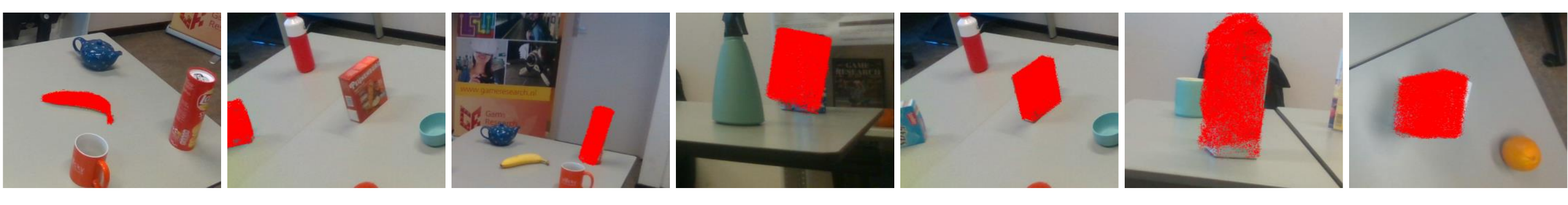}                %
	\end{minipage}}	
	\caption{Examples of accuracy performance. Each 3D model is projected to the image plane with the estimated 6D pose. }
	\label{Fig.8}                                                        %
\end{figure*}

 Furthermore, we also visualize the comparison results  as shown in Figure \ref{Fig.8}.  It can be seen that  DenseFusion, ASS3D and GraphFusion provide more accurate 6D pose for colorful objects, such as banana, gingerbread box and  chips can, while these approaches are less  robust against   dark color or low texture objects, such as cookie box and milk box.

\section{Conclusion}
	
The 6D object pose estimation  is a challenging but important research direction for virtual reality, robotics and visual navigation. With this benchmark, we have captured some state-of-the-art approaches in this field and will be able to systematically measure its progress in the future. The evaluation results indicate  that  the approach fully exploiting color and depth features performs best, outperforming pixel fusion based method and the approach  with multimodal supervision. As open problems, our analysis takes varying  texture and shape objects,   and  object symmetries  into consideration. We also note some limitations of our datasets, which we hope to improve in the future. Firstly, the synthetic dataset needs to be expanded by adding more reflective objects, occlusion, varying lighting conditions and objects with different sizes. On the other hand, more accurate depth maps and 3D models  need to be provided.

\section{Acknowledgements}
Part of ASS3D research has been supported by the European Commission funded program FASTER, under H2020 Grant Agreement 833507. The organizers    have been partially supported by the China Scholarship Council (CSC).

\bibliographystyle{eg-alpha-doi} 
\bibliography{shrec}

\newcommand{\etalchar}[1]{$^{#1}$}
\begin{thebibliography}{\uppercase{MAMT15}}

\bibitem[AZD{\etalchar{*}}20]{Albanis_2020_ECCV_Workshops}
\textsc{Albanis G., Zioulis N., Dimou A., Zarpalas D., Daras P.}:
\newblock Dronepose: Photorealistic uav-assistant dataset synthesis for 3d pose
  estimation via a smooth silhouette loss.
\newblock In \emph{Proceedings of the European Conference on Computer Vision
  (ECCV) Workshops} (August 2020).

\bibitem[BKM{\etalchar{*}}14]{brachmann2014learning}
\textsc{Brachmann E., Krull A., Michel F., Gumhold S., Shotton J., Rother C.}:
\newblock Learning 6d object pose estimation using 3d object coordinates.
\newblock In \emph{European conference on computer vision} (2014), Springer,
  pp.~536--551.

\bibitem[CC16]{choi2016rgb}
\textsc{Choi C., Christensen H.~I.}:
\newblock Rgb-d object pose estimation in unstructured environments.
\newblock \emph{Robotics and Autonomous Systems 75} (2016), 595--613.

\bibitem[CDLN07]{crete2007blur}
\textsc{Crete F., Dolmiere T., Ladret P., Nicolas M.}:
\newblock The blur effect: perception and estimation with a new no-reference
  perceptual blur metric.
\newblock In \emph{Human vision and electronic imaging XII} (2007), vol.~6492,
  International Society for Optics and Photonics, p.~64920I.

\bibitem[CZA{\etalchar{*}}19]{cheng20196d}
\textsc{Cheng Y., Zhu H., Acar C., Jing W., Wu Y., Li L., Tan C., Lim J.-H.}:
\newblock 6d pose estimation with correlation fusion.
\newblock \emph{arXiv preprint arXiv:1909.12936} (2019).

\bibitem[DFI{\etalchar{*}}15]{dosovitskiy2015flownet}
\textsc{Dosovitskiy A., Fischer P., Ilg E., Hausser P., Hazirbas C., Golkov V.,
  Van Der~Smagt P., Cremers D., Brox T.}:
\newblock Flownet: Learning optical flow with convolutional networks.
\newblock In \emph{Proceedings of the IEEE international conference on computer
  vision} (2015), pp.~2758--2766.

\bibitem[HHC{\etalchar{*}}11]{hinterstoisser2011multimodal}
\textsc{Hinterstoisser S., Holzer S., Cagniart C., Ilic S., Konolige K., Navab
  N., Lepetit V.}:
\newblock Multimodal templates for real-time detection of texture-less objects
  in heavily cluttered scenes.
\newblock In \emph{2011 international conference on computer vision} (2011),
  IEEE, pp.~858--865.

\bibitem[HLI{\etalchar{*}}12]{hinterstoisser2012model}
\textsc{Hinterstoisser S., Lepetit V., Ilic S., Holzer S., Bradski G., Konolige
  K., Navab N.}:
\newblock Model based training, detection and pose estimation of texture-less
  3d objects in heavily cluttered scenes.
\newblock In \emph{Asian conference on computer vision} (2012), Springer,
  pp.~548--562.

\bibitem[HMB{\etalchar{*}}18]{hodan2018bop}
\textsc{Hodan T., Michel F., Brachmann E., Kehl W., GlentBuch A., Kraft D.,
  Drost B., Vidal J., Ihrke S., Zabulis X., et~al.}:
\newblock Bop: Benchmark for 6d object pose estimation.
\newblock In \emph{Proceedings of the European Conference on Computer Vision
  (ECCV)} (2018), pp.~19--34.

\bibitem[HZRS16]{he2016deep}
\textsc{He K., Zhang X., Ren S., Sun J.}:
\newblock Deep residual learning for image recognition.
\newblock In \emph{Proceedings of the IEEE conference on computer vision and
  pattern recognition} (2016), pp.~770--778.

\bibitem[JMP{\etalchar{*}}18]{jafari2018ipose}
\textsc{Jafari O.~H., Mustikovela S.~K., Pertsch K., Brachmann E., Rother C.}:
\newblock ipose: instance-aware 6d pose estimation of partly occluded objects.
\newblock In \emph{Asian Conference on Computer Vision} (2018), Springer,
  pp.~477--492.

\bibitem[KGC15]{kendall2015posenet}
\textsc{Kendall A., Grimes M., Cipolla R.}:
\newblock Posenet: A convolutional network for real-time 6-dof camera
  relocalization.
\newblock In \emph{Proceedings of the IEEE international conference on computer
  vision} (2015), pp.~2938--2946.

\bibitem[KLS14]{kneip2014upnp}
\textsc{Kneip L., Li H., Seo Y.}:
\newblock Upnp: An optimal o(n) solution to the absolute pose problem with
  universal applicability.
\newblock In \emph{European Conference on Computer Vision} (2014), Springer,
  pp.~127--142.

\bibitem[MAMT15]{mur2015orb}
\textsc{Mur-Artal R., Montiel J. M.~M., Tardos J.~D.}:
\newblock Orb-slam: a versatile and accurate monocular slam system.
\newblock \emph{IEEE transactions on robotics 31}, 5 (2015), 1147--1163.

\bibitem[MKB{\etalchar{*}}17]{michel2017global}
\textsc{Michel F., Kirillov A., Brachmann E., Krull A., Gumhold S., Savchynskyy
  B., Rother C.}:
\newblock Global hypothesis generation for 6d object pose estimation.
\newblock In \emph{Proceedings of the IEEE Conference on Computer Vision and
  Pattern Recognition} (2017), pp.~462--471.

\bibitem[NH03]{ng2003sift}
\textsc{Ng P.~C., Henikoff S.}:
\newblock Sift: Predicting amino acid changes that affect protein function.
\newblock \emph{Nucleic acids research 31}, 13 (2003), 3812--3814.

\bibitem[PZC{\etalchar{*}}17]{pavlakos20176}
\textsc{Pavlakos G., Zhou X., Chan A., Derpanis K.~G., Daniilidis K.}:
\newblock 6-dof object pose from semantic keypoints.
\newblock In \emph{2017 IEEE international conference on robotics and
  automation (ICRA)} (2017), IEEE, pp.~2011--2018.

\bibitem[QSMG17]{qi2017pointnet}
\textsc{Qi C.~R., Su H., Mo K., Guibas L.~J.}:
\newblock Pointnet: Deep learning on point sets for 3d classification and
  segmentation.
\newblock In \emph{Proceedings of the IEEE conference on computer vision and
  pattern recognition} (2017), pp.~652--660.

\bibitem[SF16]{schonberger2016structure}
\textsc{Schonberger J.~L., Frahm J.-M.}:
\newblock Structure-from-motion revisited.
\newblock In \emph{Proceedings of the IEEE Conference on Computer Vision and
  Pattern Recognition} (2016), pp.~4104--4113.

\bibitem[TSF18]{tekin2018real}
\textsc{Tekin B., Sinha S.~N., Fua P.}:
\newblock Real-time seamless single shot 6d object pose prediction.
\newblock In \emph{Proceedings of the IEEE Conference on Computer Vision and
  Pattern Recognition} (2018), pp.~292--301.

\bibitem[TTS{\etalchar{*}}18]{tremblay2018deep}
\textsc{Tremblay J., To T., Sundaralingam B., Xiang Y., Fox D., Birchfield S.}:
\newblock Deep object pose estimation for semantic robotic grasping of
  household objects.
\newblock \emph{arXiv preprint arXiv:1809.10790} (2018).

\bibitem[UKA{\etalchar{*}}11]{ulbrich2011opengrasp}
\textsc{Ulbrich S., Kappler D., Asfour T., Vahrenkamp N., Bierbaum A.,
  Przybylski M., Dillmann R.}:
\newblock The opengrasp benchmarking suite: An environment for the comparative
  analysis of grasping and dexterous manipulation.
\newblock In \emph{2011 IEEE/RSJ International Conference on Intelligent Robots
  and Systems} (2011), IEEE, pp.~1761--1767.

\bibitem[VCC{\etalchar{*}}17]{velivckovic2017graph}
\textsc{Veli{\v{c}}kovi{\'c} P., Cucurull G., Casanova A., Romero A., Lio P.,
  Bengio Y.}:
\newblock Graph attention networks.
\newblock \emph{arXiv preprint arXiv:1710.10903} (2017).

\bibitem[WXZ{\etalchar{*}}19]{wang2019densefusion}
\textsc{Wang C., Xu D., Zhu Y., Mart{\'\i}n-Mart{\'\i}n R., Lu C., Fei-Fei L.,
  Savarese S.}:
\newblock Densefusion: 6d object pose estimation by iterative dense fusion.
\newblock In \emph{Proceedings of the IEEE Conference on Computer Vision and
  Pattern Recognition} (2019), pp.~3343--3352.

\bibitem[XSNF17]{xiang2017posecnn}
\textsc{Xiang Y., Schmidt T., Narayanan V., Fox D.}:
\newblock Posecnn: A convolutional neural network for 6d object pose estimation
  in cluttered scenes.
\newblock \emph{arXiv preprint arXiv:1711.00199} (2017).

\bibitem[YSW{\etalchar{*}}19]{yifan2019differentiable}
\textsc{Yifan W., Serena F., Wu S., {\"O}ztireli C., Sorkine-Hornung O.}:
\newblock Differentiable surface splatting for point-based geometry processing.
\newblock \emph{ACM Transactions on Graphics (TOG) 38}, 6 (2019), 1--14.

\bibitem[ZC17]{zhang2017texture}
\textsc{Zhang H., Cao Q.}:
\newblock Texture-less object detection and 6d pose estimation in rgb-d images.
\newblock \emph{Robotics and Autonomous Systems 95} (2017), 64--79.

\end{thebibliography}

\end{document}